\pdfoutput=1
\documentclass{article}

\usepackage{microtype}
\usepackage{graphicx}
\usepackage{booktabs} %
\usepackage{hyperref}

\usepackage[accepted]{icml2024}

\usepackage{amsmath}
\usepackage{amssymb}
\usepackage{mathtools}
\usepackage{amsthm}

\usepackage[capitalize,noabbrev]{cleveref}

\theoremstyle{plain}
\newtheorem{theorem}{Theorem}[section]

\theoremstyle{definition}

\theoremstyle{remark}

\usepackage[disable,textsize=tiny]{todonotes}

\icmltitlerunning{Adaptive Hierarchical Certification for Segmentation using Randomized Smoothing}

\usepackage{amsmath}
\usepackage{amsthm}
\usepackage{amssymb}
\usepackage{bbm}
\usepackage{mathtools}

\newtheorem{prop}{Proposition}
\usepackage{multirow}
\usepackage{cleveref}
\usepackage{pgf}
\usepackage{booktabs}
\usepackage{tabularx}
\definecolor{ForestGreen}{RGB}{34,139,34}
\usepackage{dsfont}
\usepackage{amsmath}
\DeclareMathOperator*{\argmax}{arg\,max}
\DeclareMathOperator*{\argmin}{arg\,min}
\usepackage{hyperref}
\usepackage{url}
\usepackage{tikz}
\usepackage{tikz-cd}
\usepackage{pgfplots}
\pgfplotsset{compat=1.14}
\usepackage{multirow}
\usetikzlibrary{decorations.pathreplacing,calc}
\newcommand{\tikzmark}[1]{\tikz[overlay,remember picture] \node (#1) {};}
\usepackage{wrapfig}
\newcommand*{\AddNote}[4]{%
    \begin{tikzpicture}[overlay, remember picture]
        \draw [decoration={brace,amplitude=0.5em},decorate,ultra thick,black]
            ($(#3)!(#1.north)!($(#3)-(0,1)$)$) --  
            ($(#3)!(#2.south)!($(#3)-(0,1)$)$)
                node [align=center, text width=2.5cm, pos=0.5, anchor=west] {#4};
    \end{tikzpicture}
}%

\newcommand{\class}[1]{\textit{#1}}
\newcommand{\cig}{\mathrm{CIG}}

\newcommand{\probP}{\mathds{P}}
\newcommand{\myparagraph}[1]{\vspace{3pt}\noindent{\bf #1}}
\newcommand{\bernt}[1]{\textcolor[rgb]{0.08, 0.38, 0.74}{\textbf{Bernt:} #1}}
\newcommand{\mario}[1]{\textcolor{green}{Mario: #1}}
\newcommand{\tobias}[1]{\textcolor{cyan}{Tobias: #1}}
\newcommand{\alaa}[1]{\textcolor{red}{Alaa: #1}}
\newcommand{\resolved}[1]{}
\usepackage{stfloats}
\usepackage{subcaption} %

\renewcommand{\ldots}{\texttt{.\kern-0.3em.\kern-0.3em.}}
\begin{document}

\twocolumn[
\icmltitle{Adaptive Hierarchical Certification for Segmentation\\ using Randomized Smoothing}

\icmlsetsymbol{equal}{*}

\begin{icmlauthorlist}
\icmlauthor{Alaa Anani}{xxx,yyy}
\icmlauthor{Tobias Lorenz}{xxx}
\icmlauthor{Bernt Schiele}{yyy}
\icmlauthor{Mario Fritz}{xxx}
\end{icmlauthorlist}

\icmlaffiliation{yyy}{Max Planck Institute for Informatics, Saarland Informatics Campus, Saarbr\"ucken, Germany}
\icmlaffiliation{xxx}{CISPA Helmhotz Center for Information Security, Saarbr\"ucken, Germany}

\icmlcorrespondingauthor{Alaa Anani}{aanani@mpi-inf.mpg.de}
\icmlcorrespondingauthor{Tobias Lorenz}{tobias.lorenz@cispa.de}

\icmlkeywords{Machine Learning, Certification, Heirarchical certification, adaptive certification, ICML}

\vskip 0.3in
]

\printAffiliationsAndNotice{}  %

\begin{abstract}

Certification for machine learning is proving that no adversarial sample can evade a model within a range under certain conditions, a necessity for safety-critical domains. Common certification methods for segmentation use a flat set of fine-grained classes, leading to high abstain rates due to model uncertainty across many classes. We propose a novel, more practical setting, which certifies pixels within a multi-level hierarchy, and adaptively relaxes the certification to a coarser level for unstable components classic methods would abstain from, effectively lowering the abstain rate whilst providing more certified semantically meaningful information. We mathematically formulate the problem setup, introduce an adaptive hierarchical certification algorithm and prove the correctness of its guarantees. Since certified accuracy does not take the loss of information into account for coarser classes, we introduce the Certified Information Gain ($\mathrm{CIG}$) metric, which is proportional to the class granularity level. Our extensive experiments on the datasets Cityscapes, PASCAL-Context, ACDC and COCO-Stuff demonstrate that our adaptive algorithm achieves a higher $\mathrm{CIG}$ and lower abstain rate compared to the current state-of-the-art certification method. Our code can be found here: \href{https://github.com/AlaaAnani/adaptive-certify}{https://github.com/AlaaAnani/adaptive-certify}.

\end{abstract}

\section{Introduction}

\resolved{\mario{bibliography is a mess ;) . Please clean up. consistency! well known conference can use abbreviations e.g. ICML. unknown venues have to be written out. on URLS. one venue -- one name}\alaa{Tobias is working on it.}}
Image semantic segmentation is of paramount importance to many safety-critical applications such as autonomous driving \citep{kaymak2019brief, Zhang_2016_CVPR},  medical imaging \citep{kayalibay2017cnn, guo2019deep}, video surveillance \citep{cao2020ship}, and object detection \citep{Gidaris_2015_ICCV}. However, ever since deep neural networks were shown to be inherently non-robust in the face of small adversarial perturbations \citep{szegedy2013intriguing}, the risk of using them in such applications has become evident. Moreover, an arms race between new adversarial attacks and defenses has developed, which calls for the need for provably and certifiably robust defenses. Many certification techniques have been explored in the case of classification \citep{li2023sok}, with the first recent effort in semantic segmentation by \citet{fischer}.

Certification for segmentation is a hard task since it requires certifying many components (i.e., pixels) simultaneously. The naive approach would be to certify each component to its top class within a radius and then take the minimum as the overall certified radius of the image. This is problematic since a single unstable component could lead to a very small radius, or even abstain due to a single abstention. The state-of-the-art certification method for segmentation, \textsc{SegCertify} \citep{fischer}, relies on randomized smoothing \citep{cohen2019certified}. It mitigates the many components issue by abstaining from unstable components and conservatively certifies the rest. While an unstable component implies that the model is not confident about a single top class, it often means that it fluctuates between classes that are semantically related (a result of our analysis in Section \ref{sec:experiments} and extended in App.~\ref{subsection:fluctuations}). For example, if an unstable component fluctuates between \class{car} and \class{truck}, certifying it within a semantic hierarchy as \class{vehicle} would provide a more meaningful guarantee compared to abstaining. 

\begin{figure*}[!ht]
\captionsetup[subfigure]{justification=centering}
\centering
\subfloat[Input image]
{\resizebox{0.3\textwidth}{!}{\includegraphics{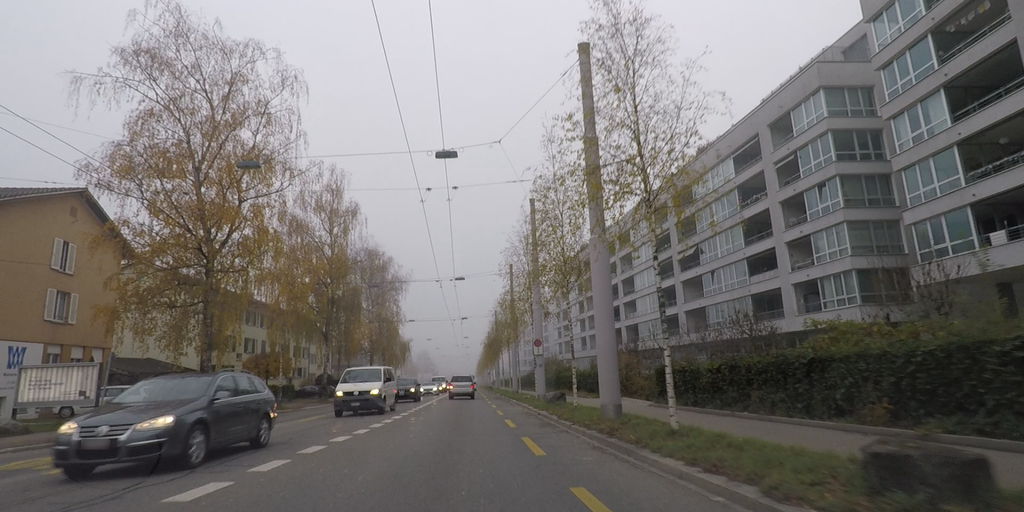}}}  
\hspace{0.5em}
\subfloat[\textsc{SegCertify} (Non-adaptive)\\
$\cig=0.76$\\
$\%\oslash=19.0$\\]  
{\resizebox{0.3\textwidth}{!}{\includegraphics{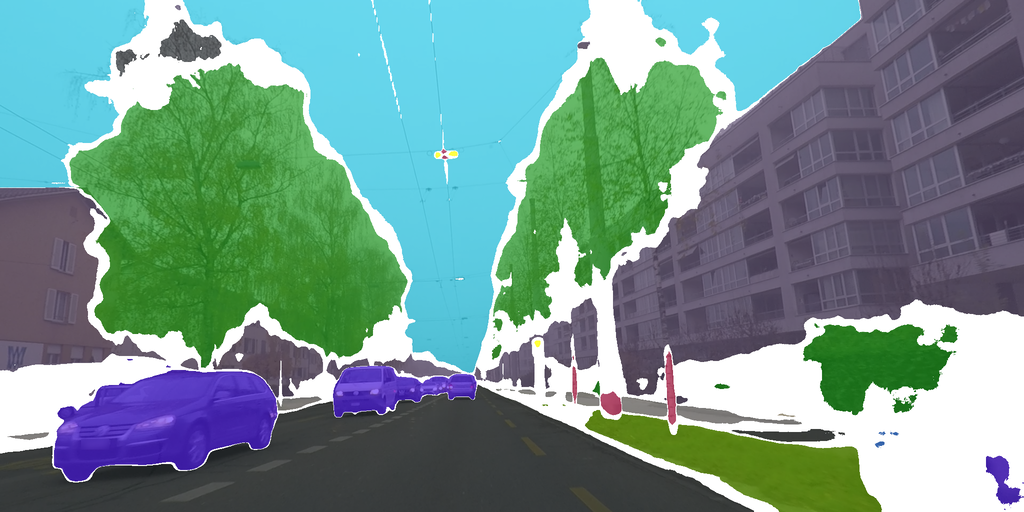}}} 
\hspace{0.5em}
\subfloat[
\textsc{AdaptiveCertify:}\\
$\cig=0.80 \textcolor{ForestGreen}{(\uparrow5.3\%)}$\\
$\%\oslash=8.2  \textcolor{ForestGreen}{(\downarrow57\%)}$]  
{\resizebox{0.3\textwidth}{!}{\includegraphics{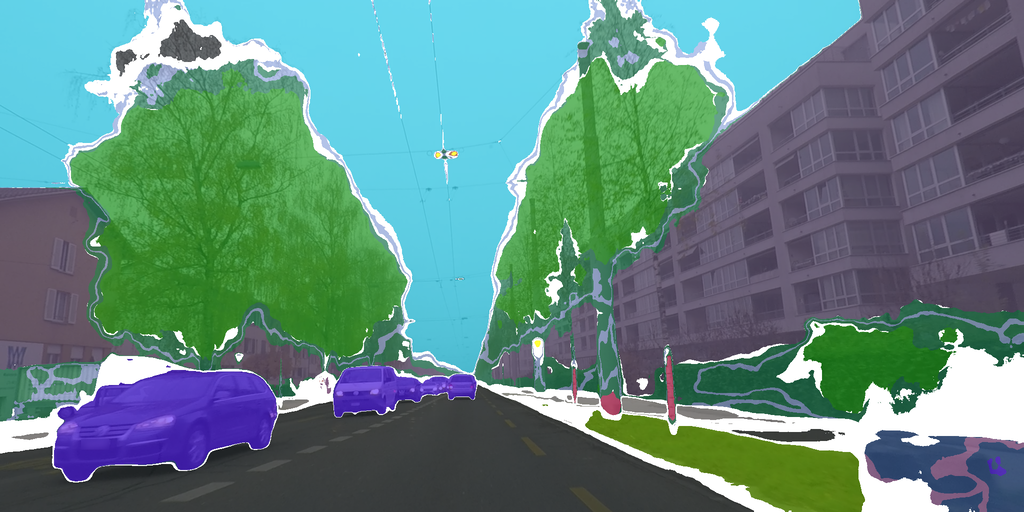}}}
\\
\subfloat[Input image]
{\resizebox{0.3\textwidth}{!}{\includegraphics{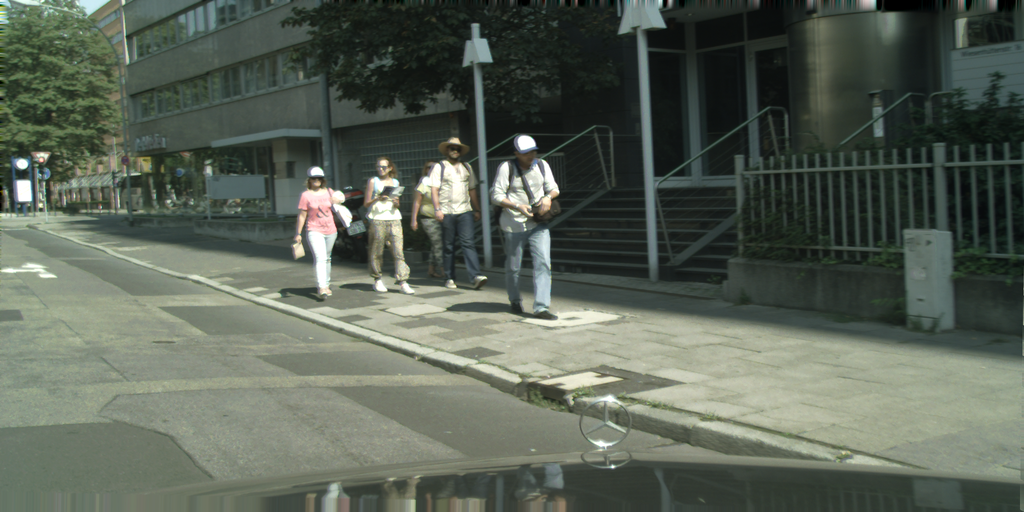}}}
\hspace{0.5em}
\subfloat[
{\textsc{SegCertify} (Non-adaptive)\\
$\cig=0.88$\\
$\%\oslash=9.0$\\}]
{\resizebox{0.3\textwidth}{!}{\includegraphics{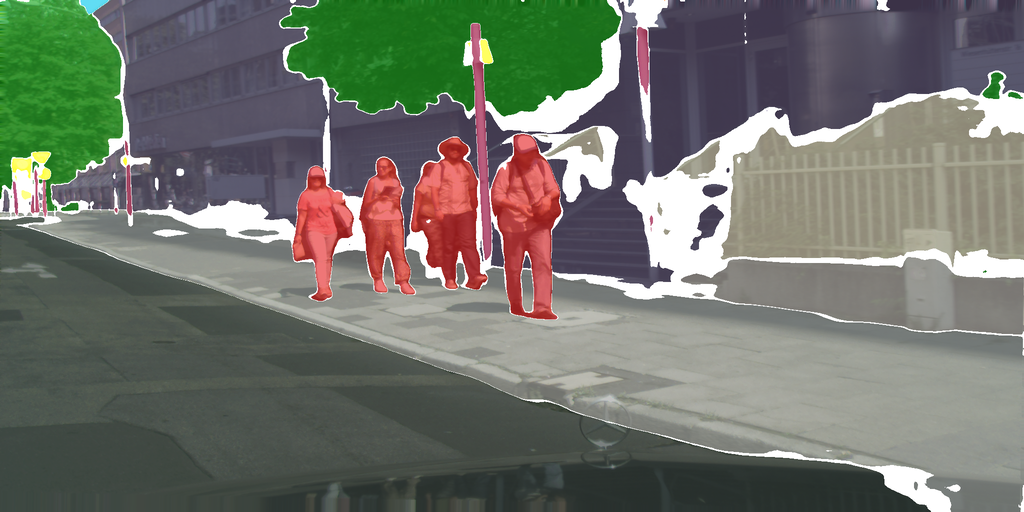}}}
\hspace{0.5em}
\subfloat[
\textsc{AdaptiveCertify:}\\
$\cig=0.90 \textcolor{ForestGreen}{(\uparrow2.3\%)}$\\
$\%\oslash=3.3\textcolor{ForestGreen}{(\downarrow63\%)}$\\
]
{\resizebox{0.3\textwidth}{!}{\includegraphics{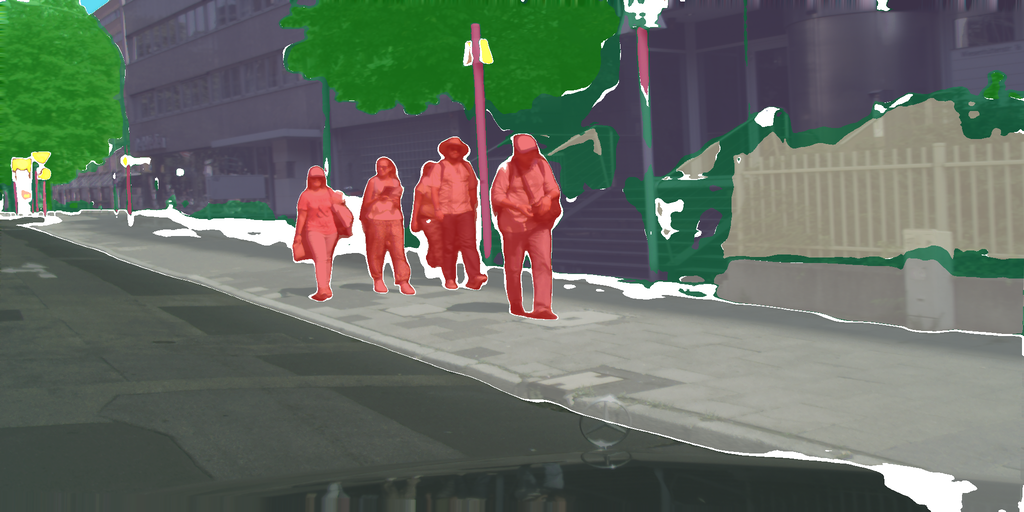}}}

\caption{The certified segmentation outputs on input images (a) and (d) from \textsc{SegCertify} in (b) and (e), and \textsc{AdaptiveCertify} in (c) and (f) with their corresponding Certified Information Gain ($\cig$) and abstain rate \%$\oslash$. Our method provides more meaningful certified output in pixels the state-of-the-art abstains from (white pixels), with a much lower abstain rate, and higher $\cig$. The segmentation color palette can be found in Figure \ref{fig:cs-h}.
\resolved{\tobias{Would it make sense to have a legend somewhere what color maps to which classes? We could also highlight some of the improved areas, either visually or in text.}\alaa{The color palette is in the big hierarchy graph in the appendix. I also referenced it the caption here.}}}
\label{fig:teaser}
    \end{figure*}

\resolved{\tobias{this last sentence needs some more explanation. It's the key problem we are addressing.}} \resolved{\mario{the argument with randomized smoothing is too long and complicated. definitely too long and confusing for the abstract.}}

We propose a novel hierarchical certification method for semantic segmentation, which adaptively certifies pixels within a multi-level hierarchy while preserving similar theoretical guarantees to \citet{fischer}. The hierarchy levels start from fine-grained labels to coarser ones that group them. Our algorithm relies on finding unstable components within an image and relaxing their label granularity to be certified within a coarser level in a semantic hierarchy. Meanwhile, stable components can still be certified within a fine-grained level. As depicted in Figure \ref{fig:teaser}, our approach lowers the abstain rate while providing more certified information to the end-user compared to the state-of-the-art method.

To evaluate our method, we propose a novel evaluation paradigm that accounts for the hierarchical label space, namely the Certified Information Gain (CIG). The Certified Information Gain is proportional to the granularity of the certified label; a parent vertex (e.g., \class{vehicle}) has less information gain than its children (e.g., \class{car}, \class{truck}, \class{bus}, etc.) since it provides more general information, while leaf vertices have the most information gain (i.e., the most granular classes in the hierarchy). CIG is equivalent to certified accuracy if the defined hierarchy is flat. %

\resolved{\bernt{is that measure really novel? I understood previously that this was proposed in another context and we use/adpat it for our purpose? If the latter is the case we should rephrase I suppose} \mario{information gain was used for semantic back off in classification - but Certified Information Gain is novel!}}

\textbf{Main Contributions}. Our main contributions are the following: (i) We introduce adaptive hierarchical certification for image semantic segmentation by mathematically formulating the problem and its adaptation to a pre-defined class hierarchy, (ii) We propose \textsc{AdaptiveCertify}, the first adaptive hierarchical certification algorithm, which certifies image pixels within fine-to-coarse hierarchy levels, (iii) We employ a novel evaluation paradigm for adaptive hierarchical certification: the Certified Information Gain metric and (iv) We extensively evaluate our algorithm, showing that certifying each pixel within a multi-level hierarchy achieves a significantly lower abstain rate and higher Certified Information Gain than the current state-of-the-art certification method for segmentation. Our analysis further shows the generalization of \textsc{AdaptiveCertify} with respect to different noise levels and pixel types.

\section{Related Works}

\myparagraph{Certification.}
The competition between adversarial attacks and defenses has resulted in a desire for certifiably robust approaches for verification and training \citep{li2023sok}. Certification is proving that no adversarial sample can evade the model within a guaranteed region under certain conditions \citep{papernot2018sok}. There are two major lines of certifiers, deterministic and probabilistic techniques. 

Deterministic certification techniques such as SMT solvers \citep{pulina2010abstraction, pulina2012challenging}, Mixed-Integer Linear Programming (MILP) \citep{cheng2017maximum, dutta2018output} or Extended Simplex Method \citep{katz2017reluplex} mostly work for small networks. To certify bigger networks, an over-approximation of the network's output corresponding to input perturbations is required \citep{salman2019convex, gowal2019scalable}, which underestimates the robustness. 

Probabilistic methods work with models with added random noise: \textit{smoothed models}. Currently, only probabilistic certification methods are scalable for large datasets and networks \citep{li2023sok}. Randomized smoothing is a probabilistic approach introduced for the classification case against $l_p$ \citep{cohen2019certified} and non-$l_p$ threat models \citep{pmlr-v108-levine20a}. Beyond classification, it has been used in median output certification of regression models \citep{NEURIPS2020_0dd1bc59}, center-smoothing \citep{kumar2021center} to certify networks with a pseudo-metric output space, and most relevant to our work, scaled to certify semantic segmentation models \citep{fischer}. We expand on randomized smoothing and \citet{fischer} in Section \ref{section:randomizedsmoothing} to provide the necessary background for our work.

\myparagraph{Hierarchical Classification and Semantic Segmentation.}
Hierarchical classification categorizes components to nodes in a class taxonomy \citep{silla2011survey}, which can be a tree \citep{wu2005learning} or a Directed Acyclic Graph. In a DAG, nodes can have multiple parents; trees only allow one. Classifiers vary in hierarchy depth: some require fine-grained class prediction, namely mandatory leaf-node prediction (MLNP), while others allow classification at any level, namely non-mandatory leaf-node prediction (NMLNP) \citep{freitas2007tutorial}. One way to deal with NMLNP is to set thresholds on the posteriors to determine which hierarchy level to classify at \citep{ceci2007classifying}. 

Hierarchical classification can extend to pixel-wise segmentation. In \cite{Li_2022_CVPR}, a model for hierarchical semantic segmentation is introduced, using hierarchy during training. NMLNP is by no means standard in current semantic segmentation work, although from a practical perspective, it is useful for downstream tasks. Our certification for segmentation method follows an NMLNP approach: we can certify a pixel at a non-leaf node. Although we use hierarchy-related concepts from previous works, our main focus is on hierarchical certification for segmentation rather than hierarchical segmentation, using the input model as a black-box. 
\resolved{\bernt{not sure if we can claim that there is `no work done' -- I am almost certain that one can find some work somewhere -- but we do not need that claim imho. I would rather write something like: NMLNP is by no means standard in current semantic segmentation work, even though it is, also from a practical perspective, useful for many down-stream tasks. (or somethjing like that)}\alaa{Rephrased it.}}

\section{Preliminaries: Randomized Smoothing for Segmentation}\label{section:randomizedsmoothing}

In this section, we provide an overview of the essential background and notations needed to understand randomized smoothing for classification and segmentation, which we build on when we introduce our adaptive method.

\myparagraph{Classification.} The core idea behind randomized smoothing \citep{cohen2019certified} is to construct a smoothed classifier $g$ from a base classifier $f$. The smoothed classifier $g$ returns the class the base classifier $f$ would return after adding isotropic Gaussian noise to the input $x$. The smooth classifier is certifiably robust to $\ell_2$-perturbations within a radius $R$. Formally, given a classifier $f: \mathbb{R}^m \to \mathcal{Y}$ and Gaussian noise $\epsilon \sim \mathcal{N}(0, \sigma^2 I)$, the smoothed classifier $g: \mathbb{R}^m \to \mathcal{Y}$ is defined as: 
\begin{equation}
g(x) \coloneqq \argmax_{a \in \mathcal{Y}} \probP (f(x+\epsilon) = a).
\end{equation}
Then the robustness guarantee on $g$ is that it is robust to any perturbation $\delta \in \mathbb{R}^m$ added to $x$, $g(x+\delta)=g(x)$, as long as $\delta$ is $\ell_2$-bounded by the certified radius: $||\delta||_2 \leq R$. To evaluate $g$ at a given input $x$ and compute the certification radius $R$, one cannot compute $g$ exactly for black-box classifiers due to its probability component. \citet{cohen2019certified} use a Monte-Carlo sampling technique to approximate $g$ by drawing $n$ samples from $\epsilon \sim \mathcal{N}(0, \sigma^2 I)$, evaluating $f(x+\epsilon)$ at each, and then using its outputs to estimate the top class and certification radius with a pre-set confidence of $1-\alpha$, such that $\alpha$ $\in [0, 1)$ is the type I error probability.

\paragraph{Segmentation.}

To adapt randomized smoothing to segmentation, \citet{fischer} propose a mathematical formulation for the problem and introduce the scalable \textsc{SegCertify} algorithm to certify any segmentation model $f: \mathbb{R}^{N\times m} \to \mathcal{Y}^N$, such that $N$ is the number of components (i.e., pixels), and $\mathcal{Y}$ is the classes set. The direct application of randomized smoothing is done by applying the certification component-wise on the image. This is problematic since it gets affected dramatically by a single bad component by reporting a small radius or abstaining from certifying all components. \textsc{SegCertify} circumvents the bad components issue by introducing a strict smooth segmentation model, that abstains from a component if the top class probability is lower than a threshold $\tau \in [0, 1)$. The smooth model $g^{\tau}: \mathbb{R}^{N\times m} \to \hat{\mathcal{Y}}^N$ is defined as:
\begin{equation}
g^{\tau}_i(x) = 
    \begin{cases}
        c_{A,i} & \text{if } \mathds{P}_{\epsilon \sim \mathcal{N}(0, \sigma^2 I)}(f_i(x+\epsilon)) >\tau, \\
        \oslash & \text{otherwise} 
    \end{cases}
    \label{eq:segcertify}
\end{equation}
where $c_{A,i} = \argmax_{c\in\mathcal{Y}} \probP_{\epsilon \sim \mathcal{N}(0, \sigma^2 I)}(f(x+\epsilon) = c)$ and  $\hat{\mathcal{Y}} = \mathcal{Y} \cup \{\oslash\}$ is the set of class labels combined with the abstain label. For all components where $g^{\tau}_i(x)$ commits to a class (does not abstain), the following theoretical guarantee holds:

\begin{theorem}[from \citep{fischer}]
Let $\mathcal{I}_x=\{i \mid g_i^{\tau} \neq \oslash, i\in 1, \ldots, N\}$ be the set of certified components indices in $x$. Then, for a perturbation $\delta \in \mathbb{R}^{N\times m}$ with $||\delta||_2 < R \coloneqq \sigma \Phi^{-1}(\tau)$, for all $i\in \mathcal{I}_x$: $g_i^{\tau}(x+\delta) = g_i^{\tau}(x)$.
\label{thm:cert}
\end{theorem}

That is, if the smoothed model $g_i^\tau$ commits to a class, then it is certified with a confidence of $1-\alpha$ to not change its output $g_i^\tau(x) = g_i^\tau(x+\delta)$ for all perturbations that are $\ell_2$-bounded by the certified radius:  $\forall_\delta: ||\delta||_2 \leq R$.

To estimate $g^{\tau}$, \citet{fischer} employ a Monte-Carlo sampling technique in \textsc{SegCertify} to draw $n$ samples from $f(x+\epsilon)$ where $\epsilon \sim \mathcal{N}(0, \sigma^2 I)$, while keeping track of the class frequencies per pixel. With these frequencies, $p$-values are computed for hypothesis testing that either result in certification or abstain. Since there are $N$ tests performed at once, a multiple hypothesis scheme is used to bound the probability of type I error (family-wise error rate) to $\alpha$.

\section{Adaptive Hierarchical Certification}\label{section:method}
 Abstaining from all components with a top-class probability $\leq \tau$ as previously described is a conservative requirement. While it mitigates the bad components effect on the certification radius by abstaining from them, those components are not necessarily ``bad'' in principle. Bad components have fluctuating classes due to the noise during sampling, which causes their null hypothesis to be accepted, and hence, are assigned $\oslash$. While this is a sign of the lack of the model's confidence in a single top class, it often means that the fluctuating classes are semantically related (a result of our analysis in Section \ref{sec:experiments} and App.~\ref{subsection:fluctuations}). For example, if sampled classes fluctuate between \class{rider} and \class{person}, this is semantically meaningful and can be certified under a label \class{human} instead of abstaining. This motivates the intuition behind our hierarchical certification approach, which relaxes the sampling process to account for the existence of a hierarchy.
 
\begin{figure*}[b]
\centering
\resizebox{\textwidth}{!}{\includegraphics[]{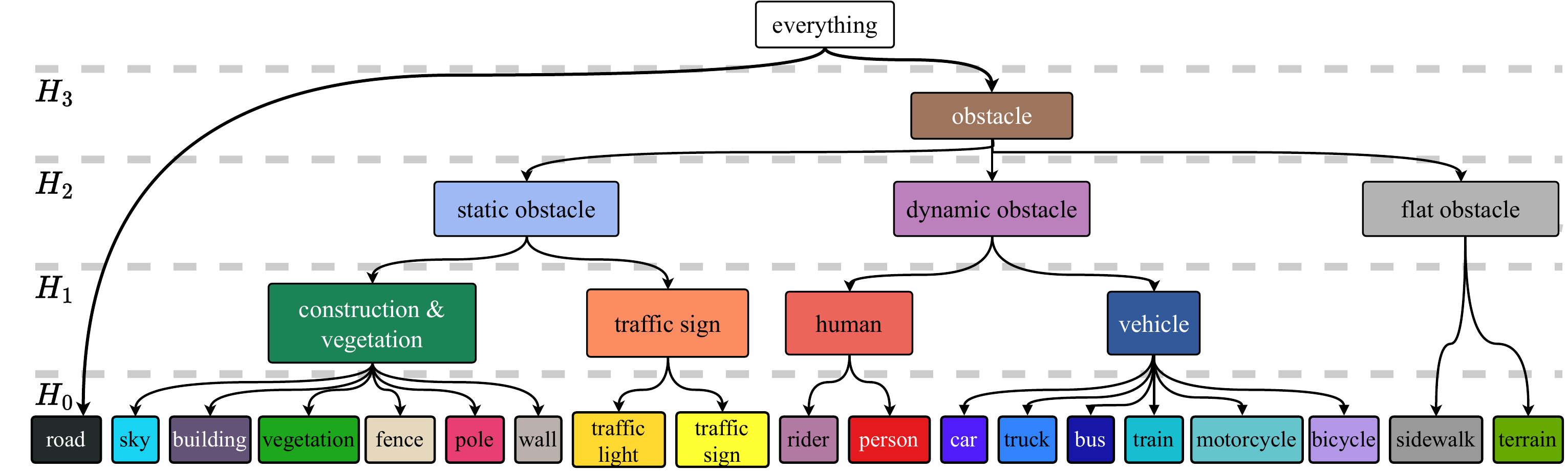}}  
\caption{A DAG representing a semantic hierarchy on top of the $19$ classes of Cityscapes \citep{cityscapes}. The node colors represent the color palette used in the segmentation results. Hierarchies on all datasets are described in App.~\ref{subsection:datasets}.}
\setlength{\belowcaptionskip}{-1pt} %
\label{fig:cs-h}
\end{figure*}

\myparagraph{Challenges:} To construct a certifier that adaptively groups the fluctuating components' outputs, there are three challenges to solve: (i) \textbf{Finding fluctuating components:} The question is: how do we find fluctuating or unstable components? Using the samples that are used in the statistical test would violate it since the final certificate should be drawn from i.i.d samples, (ii) \textbf{Adaptive sampling:} Assuming fluctuating components were defined, the adjustment of the sampling process to group semantically similar labels while working with a flat base model can be tricky. The challenge is to transform a model with flat, fine-grained labels into one whose output labels are part of a hierarchy while dealing with said model as a black-box, and (iii) \textbf{Evaluation:} Given a certifier that allows a component to commit to coarser classes, we need a fair comparison to other classical flat-hierarchy certification approaches (e.g., \textsc{SegCertify}). It is not fair to use the certified accuracy since it does not account for the information loss in croarser classes.

We construct a generalization of the smoothed model which operates on a flat-hierarchy of fine-grained classes in Eq. \ref{eq:segcertify} to formulate a hierarchical version of it. To recall in the definition, a smooth model $g^\tau$ certifies a component if it commits to a top class whose probability is $ > \tau$, otherwise it abstains. The construction of $g^\tau$ deals with the model $f$ as a black-box, that is, by plugging in any different version of $f$, the same guarantees in Theorem \ref{thm:cert} hold. We show the mathematical formulation of how we construct a hierarchical version of the smoothed model, and discuss how we overcome the challenges associated with it in this section.

\subsection{Hierarchical Certification: Formulation}\label{sec:formulation}
To define a hierarchical version of the smoothed model, we first replace the flat-hierarchy set of classes $\mathcal{Y}$ with a pre-defined class hierarchy graph $H=(\mathcal{V}, \mathcal{E})$, where the vertex set $\mathcal{V}$ contains semantic classes and the edge set $\mathcal{E}$ contains the relation on the vertices. Second, we define a hierarchical version of $f$, namely $f^{H}: \mathbb{R}^{N\times m} \to \mathcal{V}^N$, that maps the image components (pixels) with $m$ channels ($m=3$ for RGB) to the vertices $\mathcal{V}$. Third, we define a hierarchical smoothed model $c^{\tau, H}: \mathbb{R}^{N\times m} \to \hat{\mathcal{V}}^N$, such that $\hat{\mathcal{V}} = \mathcal{V} \cup \{\oslash\}$:
\begin{equation}\label{eq:c}
\textcolor{blue}{c^{\tau, H}}_i(x) = 
    \begin{cases}
        \textcolor{blue}{v_{A,i}} & \text{if } \probP_{\epsilon \sim \mathcal{N}(0, \sigma^2 I)}(\textcolor{blue}{f^{H}_i}(x+\epsilon)) >\tau \\
        \oslash, & \text{otherwise} 
    \end{cases}
\end{equation}
where $v_{A,i} = \argmax_{v \in \mathcal{V}} \probP_{\epsilon \sim \mathcal{N}(0, \sigma^2 I)}(f^{H}_i(x+\epsilon) =v)$. This certifier $c^{\tau, H}$ has three main novel components: the hierarchy graph $H$ (Section \ref{sec:hierarchy}), the hierarchical function $f^{H}$ (Section \ref{sec:fhie}) and the certification algorithm to compute $c^{\tau, H}$ (Section \ref{sec:algo}).
\resolved{\bernt{detail: would it make sense to already highlight these differences in the equation above e.g. make these symbols bold or in color? The reasons is that the equation looks very similar to equation (2) and thus feels quite incremental at best}}

\subsection{The Class Hierarchy Graph}\label{sec:hierarchy}

We design the class hierarchy $H$ used by $c^{\tau, H}$ to capture the semantic relationship amongst the classes in $\mathcal{Y}$, as illustrated in the hierarchy we build on top of the 19 classes of Cityscapes in Figure \ref{fig:cs-h}. The full hierarchies on all other datasets are described in App.~\ref{subsection:datasets}. $H$ is a pair $(\mathcal{V}, \mathcal{E})$ representing a DAG, where $\mathcal{V}$ is the set of vertices, and $\mathcal{E}$ is the set of edges representing the \textsc{IS-A} relationship among edges. We do not allow more than one parent for each vertex. An edge $e=(u, v)\in \mathcal{E}$ is defined as a pair of vertices $u$ and $v$, which entails that $u$ is a parent of $v$. The root vertex of the DAG denotes the most general class, \class{everything}, which we do not use. The hierarchy is divided into multiple levels $H_0, \ldots, H_L$, the more fine-grained the classes are, the lower the level. A hierarchy level is a set $H_l$ of the vertices falling within it. Leaf vertices $\mathcal{Y}$ are not parents of any other vertices. Essentially, $\mathcal{Y} =  H_0$.

\subsection{Fluctuating Components and Adaptive Sampling}\label{sec:fhie}
In this part, we discuss how to solve two of the challenges concerning constructing an adaptive hierarchical certifier: defining fluctuating components without using the samples in the statistical test, and the adjustment of the sampling process to be adaptive. 

We define the fluctuating components by an independent set of samples from those used in the hypothesis test. We first draw initial $n_0$ posterior samples per component from the segmentation head of $f$, defined as $f_\mathrm{seg}: \mathbb{R}^{N\times m} \to [0, 1]^{N \times |\mathcal{Y}|}$. We then look at the top two classes' mean posterior difference. The smaller the difference, the coarser the hierarchy level the component is assigned to. These steps are outlined in Algorithm \ref{alg:component-to-level} describing \textsc{GetComponentLevels}, which finds the hierarchy level index for every component. 
\resolved{\tobias{Single / few words at the end of paragraphs are good places to look for saving space. The previous paragraph can almost always be changed to take less space, e.g., by removing filler words.}\alaa{I rephrased it. Thank you.}}

We invoke \textsc{SamplePosteriors} to draw initial $n_0$ samples from $f_\mathrm{seg}(x+\epsilon)$ with $\epsilon \sim \mathcal{N}(0, \sigma^2 I)$. This method retrieves $n_0$ posteriors per component: ${Ps}^0_1, \ldots, {Ps}^0_N$, such that ${Ps}^0_i$ is a set of $n_0$ posterior vectors $\in [0, 1]^{|\mathcal{Y}|}$ for the $i^{\text{th}}$ component, outlined in App. Algorithm~\ref{alg:sample-posteriors}. Then, for every component $i$, we get the mean of its $n_0$ posteriors ${P}^0_i$, and calculate the posterior difference $\Delta P_i$ between the top two classes, indexed by $\hat{c}_{A_i}$ and $\hat{c}_{B_i}$. We use thresholds to determine its hierarchy level index $l$ by invoking a threshold function $T_{\mathrm{thresh}}$. Given a hierarchy with $L$ levels, the threshold function $T_{\mathrm{thresh}}$ is defined as:
\begin{equation}\label{eq:fthresh}
    T_{\mathrm{thresh}}(\Delta P_i) = \argmin_{l\in \{0, \ldots , L-1\}} t_l, \text{ s.t. } t_l < \Delta P_i
\end{equation}
with $t_l\in [0, 1]$. $T_{\mathrm{thresh}}$ returns the index of the most fine-grained hierarchy level the component can be assigned to based on the pre-set thresholds $t_0 > t_1 > \ldots > t_{L-1}$.

\begin{algorithm}[htbp]
\caption{\textsc{GetComponentLevels}: algorithm to map components to hierarchy levels}\label{alg:component-to-level}
\begin{algorithmic}
\FUNCTION{\textsc{GetComponentLevels}($f$, $x$, $n_0$, $\sigma$)}

\STATE $Ps^0_1, \ldots, Ps^0_N \gets$ \textsc{SamplePosteriors}($f$, $x$, $n_0$, $\sigma$) 

      \FOR{i $\gets$ {1, \ldots, N}}
        \STATE $P^0_i \gets $ \text{mean} $Ps^0_i$
        \STATE $\hat{c}_{A_i}, \hat{c}_{B_i} \gets $ \text{top two class indices} $P^0_i$
        \STATE $\Delta P_i \gets   P^0_i[\hat{c}_{A_i}] - P^0_i[\hat{c}_{B_i}]$
        \STATE $l_i \gets T_{\mathrm{thresh}}(\Delta P_i )$
      \ENDFOR

\STATE \textbf{return} ($l_1, \ldots, l_N$), ($\hat{c}_{A_1}, \ldots, \hat{c}_{A_N}$)
\ENDFUNCTION
\end{algorithmic}
\end{algorithm}

Now that we know which level index $l_i$ every component is mapped to, we can define $f^{H}$, which takes an image $x$ and does a pixel-wise mapping to vertices $\hat{v}$ within every component's assigned hierarchy level $H_{l_i}$. Mathematically, we define the predicted label $\hat{v}_i$ for component $i$ as:
\begin{equation} \label{eq:fh}
\begin{split}
&{f^{{H}}_i}(x) = K(f_i(x), l_i) =  \hat{v}_i \iff\\
&\exists_{\hat{v}_i,u_1, \ldots ,u_p,\hat{y}_i} (\{(\hat{v}_i, u_1),\ldots,(u_p, \hat{y}_i)\} \subseteq \mathcal{E}) \wedge (\hat{v}_i \in H_{l_i})
\end{split}
\end{equation}
such that there is a path from the parent vertex $\hat{v}_i$ that belongs to the hierarchy level $H_{l_i}$ to the predicted leaf $\hat{y}_i=f_i(x)$. For example, using the hierarchy Figure \ref{fig:cs-h}, $K(\class{bus}, 0) = \class{bus}, K(\class{bus}, 1)=\class{vehicle}$ and $ K(\class{bus}, 2) = \class{dynamic obstacle}$. Constructing a smoothed version of $f^{H}$, namely $c_i^{\tau, H}$, is now equivalent to the hierarchical certifier we formulated earlier in Eq. \ref{eq:c}.

Evaluating $c^{\tau, H}$ requires a sampling scheme over $f^{H}$ to get the top vertex frequencies $\mathrm{cnts}_1, \ldots, \mathrm{cnts}_N$ as outlined in Algorithm \ref{alg:hsample}. The sampling of ${f}^H(x+\epsilon)$ with $\epsilon \sim \mathcal{N}(0, \sigma^2 I)$ is a form of adaptive sampling over $f$. For the $i^\text{th}$ component with level $l_i$, $f^{H}$ is invoked on $x+\epsilon$, which in its definition invokes $f(x+\epsilon)$ to output a flat segmentation label map $\hat{y}$, whose components $\hat{y}_i$ is mapped to its parent vertex $\hat{v}_i$ in $H_{l_i}$ using the function $K$ as in Eq. \ref{eq:fh}. 

\begin{algorithm}[htbp]
\caption{\textsc{HSample}: algorithm to adaptively sample}\label{alg:hsample}
\begin{algorithmic}
\FUNCTION{\textsc{HSample}($f$, $K$, $(l_1, \ldots, l_N)$, $x$, $n$, $\sigma$)}
\STATE $\mathrm{cnts}_1, \ldots, \mathrm{cnts}_N \gets$ initialize each to a zero vector of size $|\mathcal{V}|$
\STATE \text{draw random noise $\epsilon \sim \mathcal{N}(0, \sigma^2 I)$}
\FOR{$j \gets {1, \ldots, n}$}
    \STATE $\hat{y} = f(x+\epsilon)$ \tikzmark{top}
    \FOR{ $i \gets {1, \ldots, N}$} 
        \STATE $\hat{v}_i \gets K(\hat{y}[i], l[i])$ \tikzmark{bottom}\tikzmark{right}
        \STATE $\mathrm{cnts}_i[\hat{v}_i]$ += $1$
    \ENDFOR
\ENDFOR

\STATE \textbf{return} $\mathrm{cnts}_1, \ldots, \mathrm{cnts}_N$

\ENDFUNCTION
\end{algorithmic}
\AddNote{top}{right}{bottom}{mathematically equivalent to calling $\equiv f^H(x+\epsilon)$}
\end{algorithm}

\subsection{Our Algorithm: \textsc{AdaptiveCertify}}\label{sec:algo}
Putting it all together, we now introduce  \textsc{AdaptiveCertify} \ref{alg:adaptive}, which overcomes the challenges of defining the fluctuating components and employing an adaptive sampling scheme to certify an input segmentation model $f$ given a hierarchy $H$. Our certification algorithm approximates the smoothed model $c^{\tau, H}$ following a similar approach by \citet{fischer}.

\begin{algorithm}[htbp]
\caption{\textsc{AdaptiveCertify}: algorithm to hierarchically certify and predict}\label{alg:adaptive}
\begin{algorithmic}

\FUNCTION{\textsc{AdaptiveCertify}($f$, $K$, $\sigma$, $x$, $n$, $n_0$, $\tau$, $\alpha$)}

\STATE $(l_1, \ldots, l_N), (\hat{c}_{A_1}, \ldots, \hat{c}_{A_N}) \gets \textsc{GetComponent-}$\\$\textsc{Levels}$$(f, x, n_0, \sigma)$

\STATE $\hat{v}_1, \ldots, \hat{v}_N \gets $ \text{Use $K(\hat{c}_{A_i}, l_i)$ to get parent vertices of} $\hat{c}_{A_1}, \ldots, \hat{c}_{A_N}$
\STATE $\mathrm{cnts}_1, \ldots, \mathrm{cnts}_N \gets \textsc{HSample}(f, K, (l_1, \ldots, l_N),  x, $ $ n, \sigma)$

\STATE $pv_1, \ldots, pv_N \gets \textsc{BinPValue}((\hat{v}_1, \ldots, \hat{v}_N),(\mathrm{cnts}_1,$ $ \ldots, \mathrm{cnts}_N), \tau)$

\STATE $\hat{v}_1, \ldots, \hat{v}_N \gets \textsc{HypothesesTesting}(\alpha, \oslash, (pv_1,$ $\ldots,pv_N), (\hat{v}_1, \ldots, \hat{v}_N$))

\STATE $R \gets \sigma \Phi^{-1}(\tau)$
\STATE \textbf{return} $\hat{v}_1, \ldots, \hat{v}_N$, $R$
\ENDFUNCTION
\end{algorithmic}
\end{algorithm}
On a high level, \textsc{AdaptiveCertify} consists of three parts: (i) mapping components to hierarchy level indices by invoking \textsc{GetComponentLevels}, (ii) adaptively sampling to estimate the smoothed model $c^{\tau, H}$ by invoking \textsc{HSample}, and (iii) employing multiple hypothesis testing via \textsc{HypothesesTesting} (outlined in App. Algorithm~\ref{alg:h-testing}) to either certify a component or assign $\oslash$ to it. To avoid invalidating our hypotheses test, we use the initial set of $n_0$ independent samples drawn in \textsc{GetComponentLevels} to both decide on the assigned component levels indices $l_1, \ldots, l_N$, as well as the top class indices $\hat{c}_{A_1}, \ldots, \hat{c}_{A_N}$. Since those classes are in $\mathcal{Y}$ as they come from the flat model $f$, we transform them using the mapping function $K$ and the levels to get their corresponding parent vertices in the hierarchy: $\hat{v}_1, \ldots, \hat{v}_N$. These vertices are used to decide on the top vertex class, while the counts $\mathrm{cnts}_1, \ldots, \mathrm{cnts}_N$ drawn from the adaptive sampling function \textsc{HSample} are used in the hypothesis testing. With these counts, we perform a one-sided binomial test on every component to retrieve its $p$-value, assuming that the null hypothesis is that the top vertex class probability is $\leq \tau$. Then, we apply \textsc{HypothesesTesting} (App. Algorithm~\ref{alg:h-testing}) to reject (certify) or accept (abstain by overwriting $\hat{v}_i$ with $\oslash$) from components while maintaining an overall type I error probability of $\alpha$. 

We now show the soundness of \textsc{AdaptiveCertify} using Theorem \ref{thm:cert}. That is, if \textsc{AdaptiveCertify} returns a class $\hat{v_i} \neq \oslash$, then with probability $1-\alpha$, the vertex class is certified within a radius of $R:=\sigma \Phi^{-1}(\tau)$.

\begin{prop}[Similar to \citep{fischer}]
Let $\hat{v}_1,\ldots,\hat{v}_N$ be the output of \textsc{AdaptiveCertify} given an input image $x$ and $\hat{I}_x \coloneqq \{i \mid \hat{v}_i \neq \oslash\}$ be the set of non-abstain indices in $x$. Then with probability at least $1-\alpha$, $\hat{I}_x \subseteq I_x$ such that $I_x$ denotes the theoretical non-abstain indices previously defined in Theorem \ref{thm:cert} by replacing $g^{\tau}$ with our smoothed model $c^{\tau, H}$. Then, $\forall i\in \hat{I}_x$, $\hat{v}_i=c^{\tau, H}_i(x)=c^{\tau, H}_i(x+\delta)$ for  $\ell_2$-bounded noise $||\delta||_2 \leq R$.
\end{prop}
\begin{proof}
    With probability $\alpha$, a type I error would result in $i\in \hat{I}_x \setminus I_x$. However, since $\alpha$ is bounded by \textsc{HypothesesTesting}, then with probability at least $1-\alpha$, $\hat{I}_x \subseteq I_x$.
\end{proof}

\subsection{Properties of \textsc{AdaptiveCertify}}

The hierarchical nature of \textsc{AdaptiveCertify} means that instead of abstaining for unstable components, it relaxes the certificate to a coarser hierarchy level. While not always successful, this increases the chances for certification to succeed on a higher level. The abstaining can still occur on any hierarchy level, and it has two reasons: the top vertex probability is $\leq \tau$, and by definition $c^{\tau, H}$ would abstain, or it is a type II error in \textsc{AdaptiveCertify}. 

\textsc{AdaptiveCertify} guarantees that the abstain rate is always less than or equal to a non-adaptive flat-hierarchy version (e.g., \textsc{SegCertify}). If our algorithm only uses level $H_0$ for all components, the abstain rate will be equal to a non-adaptive version. So, since some components are assigned to a coarser level, their $p$-values can only decrease, which can only decrease the abstain rate.

By adapting the thresholds in $T_{\mathrm{thresh}}$ and the hierarchy definition, one can influence the hierarchy levels assigned to the components. Strict thresholds or coarser hierarchies would allow most components to fall within coarse levels. This is a parameterized part of our algorithm that can be adjusted based on the application preferences, trading off the certification rate versus the Certified Information Gain. We explore the tradeoff in App.~\ref{subsection:tradeoff}.
\resolved{\tobias{broken reference}. \alaa{We decided to remove the tradeoff graphs, I commented the reference to it.}}

\subsection{Evaluation Paradigm: Certified Information Gain} \label{sec:cig}
\resolved{\mario{I've changed the subsection to Certified Information Gain; is it still possible to consistently talk about this? and also use the abbreviation? there should also be an equation introducing CIG. also the tables should read "CIG" then - if we agree.}\alaa{Thank you Mario for the suggestion.}
\tobias{I like the idea, will make the required changes.}}
As mentioned previously, certified accuracy does not take the loss of information into account when traversing coarser hierarchy nodes as it would be trivial to maximize certified robustness by assigning all components to the topmost level. We therefore define a new Certified Information Gain ($\cig$) metric that is proportional to the class granularity level. That is, a pixel gets maximum $\cig$ if certified within the most fine-grained level $H_0$, and it decreases the higher the level.

Formally, given an image $x$ with predicted certified vertices $\hat{v} = c^{\tau, H}(x)$, ground truth  $y$, and hierarchy level map $L$, $\cig$ is defined as:
\begin{equation}
{
\cig(\hat{v}, y, L) =
    \frac{\sum\limits_{\hat{v}_i = K(y_i, L_i)}^{} \left(\log(|\mathcal{Y}|) - \log(G(\hat{v}_i))\right)}{N \cdot \log(|\mathcal{Y}|)}
}
\end{equation}
\resolved{\tobias{I fixed the equation scaling for now, but this is bad practice. We should try to fit it at normal font size. We could, e.g., abbreviate ``generality'' or replace the entire term with a function name that is defined elsewhere.}\alaa{I abbreviated it as $\mathrm{G}$ and mentioned that it refers to generality. I had a mathematical definition for generality before, but we found it unnecessarily complicated for the scope of the paper. I think this looks fine now.}}

$|\mathcal{Y}|$ is the number of leaves (i.e., the number of the most fine-grained classes), and $\mathrm{G}(v_i)$ returns the generality of the vertex $v_i$, which is defined as the number of leaf descendants of $v_i$ (formal definition in App. Eq.~\ref{eq:gen}). 
Assuming certification succeeds, $\cig$ is maximized when $\hat{v}$ are all leaf vertices since $\cig(\hat{v}, y, L) = \frac{\sum_{i=1}^N \left(\log(|\mathcal{Y}|) - 0 \right)}{N \cdot \log(|\mathcal{Y}|)}  =  1$ as the generality of a leaf vertex $G(\hat{v}_i)=1$. $\cig$ results in a score between 0 and 1 and reduces to certified accuracy for non-adaptive algorithms (\textsc{SegCertify}).

We also consider the class-average $\cig$, namely $c\cig$, which evaluates the performance on a per-class basis. It is defined by measuring the per-class $\cig$ for all classes in $\mathcal{Y}$ and then getting the average. Given any class $a\in\mathcal{Y}$, the per-class $\cig$ is defined as $\cig^a$:
\begin{equation}
    \cig^a(\hat{v}, y, L) = \frac{\sum\limits_{y_i=a} \cig_i(\hat{v}, y, L)}{ |\{i \mid y_i=a\}|}
\end{equation}
where $\cig_i(\hat{v}, y, L)$ denotes the $\cig$ of the $i^{\text{th}}$ component. Therefore, the class-average $\cig$ ($c\cig$) is expressed as:
\begin{equation}
    c\cig(\hat{v}, y, L) = \frac{1}{|\mathcal{Y}|} \sum_{a \in \mathcal{Y}} \cig^a(\hat{v}, y, L)
\end{equation}
$c\cig$ is a score between 0 and 1, and reduces to class-average certified accuracy for non-adaptive algorithms.

\section{Results} \label{sec:experiments}

\begin{table*}[bp]
\setlength\tabcolsep{2pt}
\begin{tabularx}{\textwidth}{lXrrlllllllll}
\toprule
\multicolumn{5}{l}{} & \multicolumn{2}{c}{Cityscapes} & \multicolumn{2}{c}{ACDC} & \multicolumn{2}{c}{PASCAL-Context} & \multicolumn{2}{c}{COCO-Stuff-10K} \\
\cmidrule{6-13}
&  & \multicolumn{1}{l}{$\sigma$} & \multicolumn{1}{l}{$R$} &  & $\cig\uparrow$ & $\%\oslash\downarrow$  & $\cig\uparrow$ & $\%\oslash\downarrow$ & $\cig\uparrow$ & $\%\oslash\downarrow$ & $\cig\uparrow$ & $\%\oslash\downarrow$\\
\midrule
& Uncertified HrNet & \multicolumn{1}{l}{-} & \multicolumn{1}{l}{-} &  & $0.90$ & $-$ & $0.61$ & $-$ & $0.58$ & $-$ &  $0.65$ & $-$ \\
\midrule
\multirow{7}{*}{\begin{tabular}[c]{@{}l@{}}$n = 100,$\\ $\tau=0.75$\end{tabular}} & \multirow{3}{*}{\textsc{SegCertify}} & $0.25$ & $0.17$ &  & $0.89$ & $7$ & $0.67$ & $21$ & $0.57$ & $20$ & $0.58$ & $20$ \\
&  & $0.33$ & $0.22$ &  & $0.81$ & $14$ & $0.57$ & $27$ & $0.46$ & $31$ & $0.52$ & $28$ \\
&  & 0.50 & 0.34 &   & $0.41$ & $26$ & $0.25$ & $26$ & $0.15$ & $41$ & $0.31$ & $45$ \\
& \multicolumn{8}{l}{} \\
& \multirow{3}{*}{\textsc{AdaptiveCertify}} & 0.25 & 0.17 &  & \textbf{0.90 {\fontsize{4}{0}\selectfont\textcolor{ForestGreen}{$1.1\%$}}} & \textbf{5 {\fontsize{4}{0}\selectfont\textcolor{ForestGreen}{$28.6\%$}}} & \textbf{0.68 {\fontsize{4}{0}\selectfont\textcolor{ForestGreen}{$1.5\%$}}} & \textbf{16 {\fontsize{4}{0}\selectfont\textcolor{ForestGreen}{$23.8\%$}}} & \textbf{0.58 {\fontsize{4}{0}\selectfont\textcolor{ForestGreen}{$1.8\%$}}} & \textbf{16 {\fontsize{4}{0}\selectfont\textcolor{ForestGreen}{$20.0\%$}}} & \textbf{0.60 {\fontsize{4}{0}\selectfont\textcolor{ForestGreen}{$3.4\%$}}} & \textbf{13 {\fontsize{4}{0}\selectfont\textcolor{ForestGreen}{$35.0\%$}}} \\
&  & 0.33 & 0.22 &  & \textbf{0.83 {\fontsize{4}{0}\selectfont\textcolor{ForestGreen}{$2.5\%$}}} & \textbf{10 {\fontsize{4}{0}\selectfont\textcolor{ForestGreen}{$28.6\%$}}} & \textbf{0.59 {\fontsize{4}{0}\selectfont\textcolor{ForestGreen}{$3.5\%$}}} & \textbf{22 {\fontsize{4}{0}\selectfont\textcolor{ForestGreen}{$18.5\%$}}} & \textbf{0.48 {\fontsize{4}{0}\selectfont\textcolor{ForestGreen}{$4.3\%$}}} & \textbf{26 {\fontsize{4}{0}\selectfont\textcolor{ForestGreen}{$16.1\%$}}} & \textbf{0.54 {\fontsize{4}{0}\selectfont\textcolor{ForestGreen}{$3.8\%$}}} & \textbf{18 {\fontsize{4}{0}\selectfont\textcolor{ForestGreen}{$35.7\%$}}} \\
&  & 0.50 & 0.34 &  & \textbf{0.44 {\fontsize{4}{0}\selectfont\textcolor{ForestGreen}{$7.3\%$}}} & \textbf{15 {\fontsize{4}{0}\selectfont\textcolor{ForestGreen}{$42.3\%$}}} & \textbf{0.27 {\fontsize{4}{0}\selectfont\textcolor{ForestGreen}{$8.0\%$}}} & \textbf{18 {\fontsize{4}{0}\selectfont\textcolor{ForestGreen}{$30.8\%$}}} & \textbf{0.16 {\fontsize{4}{0}\selectfont\textcolor{ForestGreen}{$6.7\%$}}} & \textbf{36 {\fontsize{4}{0}\selectfont\textcolor{ForestGreen}{$12.2\%$}}} & \textbf{0.35 {\fontsize{4}{0}\selectfont\textcolor{ForestGreen}{$12.9\%$}}} & \textbf{32 {\fontsize{4}{0}\selectfont\textcolor{ForestGreen}{$28.9\%$}}} \\
\midrule
\multirow{7}{*}{\begin{tabular}[c]{@{}l@{}}$n=500,$\\ $\tau=0.95$\end{tabular}} & \multirow{3}{*}{\textsc{SegCertify}} & 0.25 & 0.41 &  & $0.86$ & $12$ & $0.63$ & $29$ & $0.53$ & $30$ & $0.53$ & $31$\\
&  & 0.33 & 0.52 &  & $0.76$ & $22$ & $0.51$ & $39$ & $0.40$ & $46$ & $0.45$ & $41$\\
&  & 0.50 & 0.82 &  & $0.36$ & $39$ & $0.22$ & $39$ & $0.12$ & $59$ & $0.26$ & $61$ \\
& \multicolumn{8}{l}{} \\
& \multirow{3}{*}{\textsc{AdaptiveCertify}} & 0.25 & 0.41 &  & \textbf{0.87 {\fontsize{4}{0}\selectfont\textcolor{ForestGreen}{$1.2\%$}}} & \textbf{9 {\fontsize{4}{0}\selectfont\textcolor{ForestGreen}{$25.0\%$}}} & \textbf{0.64 {\fontsize{4}{0}\selectfont\textcolor{ForestGreen}{$1.6\%$}}} & \textbf{25 {\fontsize{4}{0}\selectfont\textcolor{ForestGreen}{$13.8\%$}}} & \textbf{0.54 {\fontsize{4}{0}\selectfont\textcolor{ForestGreen}{$1.9\%$}}} & \textbf{26 {\fontsize{4}{0}\selectfont\textcolor{ForestGreen}{$13.3\%$}}} & \textbf{0.56 {\fontsize{4}{0}\selectfont\textcolor{ForestGreen}{$5.7\%$}}} & \textbf{25 {\fontsize{4}{0}\selectfont\textcolor{ForestGreen}{$19.4\%$}}}\\
&  & 0.33 & 0.52 &  & \textbf{0.77 {\fontsize{4}{0}\selectfont\textcolor{ForestGreen}{$1.3\%$}}} & \textbf{18 {\fontsize{4}{0}\selectfont\textcolor{ForestGreen}{$18.2\%$}}} & \textbf{0.53 {\fontsize{4}{0}\selectfont\textcolor{ForestGreen}{$3.9\%$}}} & \textbf{34 {\fontsize{4}{0}\selectfont\textcolor{ForestGreen}{$12.8\%$}}} & \textbf{0.41 {\fontsize{4}{0}\selectfont\textcolor{ForestGreen}{$2.5\%$}}} & \textbf{41 {\fontsize{4}{0}\selectfont\textcolor{ForestGreen}{$10.9\%$}}} & \textbf{0.48 {\fontsize{4}{0}\selectfont\textcolor{ForestGreen}{$6.7\%$}}} & \textbf{33 {\fontsize{4}{0}\selectfont\textcolor{ForestGreen}{$19.5\%$}}}\\
&  & 0.50 & 0.82 &  & \textbf{0.40 {\fontsize{4}{0}\selectfont\textcolor{ForestGreen}{$11.1\%$}}} & \textbf{28 {\fontsize{4}{0}\selectfont\textcolor{ForestGreen}{$28.2\%$}}} & \textbf{0.24 {\fontsize{4}{0}\selectfont\textcolor{ForestGreen}{$9.1\%$}}} & \textbf{31 {\fontsize{4}{0}\selectfont\textcolor{ForestGreen}{$20.5\%$}}} & \textbf{0.13 {\fontsize{4}{0}\selectfont\textcolor{ForestGreen}{$8.3\%$}}} & \textbf{55 {\fontsize{4}{0}\selectfont\textcolor{ForestGreen}{$6.8\%$}}} & \textbf{0.29 {\fontsize{4}{0}\selectfont\textcolor{ForestGreen}{$11.5\%$}}} & \textbf{50 {\fontsize{4}{0}\selectfont\textcolor{ForestGreen}{$18.0\%$}}}\\
\bottomrule
\end{tabularx}
\caption{Certified segmentation results for 200 images from each dataset. We extend this table by including the mIoU, $c\cig$ and $c\%\oslash$ metrics per dataset in App. Tables \ref{tab:cityscapes-table}, \ref{tab:acdc-table}, \ref{tab:pascal_ctx-table} and \ref{tab:cocostuff-table} under App.~\ref{subsection:overall-performance}.}
\label{tab:table}
\end{table*}

We evaluate \textsc{AdaptiveCertify} in a series of experiments to show its performance against the current state-of-the-art, and illustrate its hierarchical nature. We use four segmentation datasets: Cityscapes \citep{cityscapes}, the Adverse Conditions Dataset with Correspondences (ACDC) \citep{acdc}, PASCAL-Context \citep{pascal} and COCO-Stuff-10K \citep{cocostuff}, which are described in App.~\ref{subsection:datasets} alongside their hierarchy graphs. 

We use HrNetV2 \citep{SunXLW19, wang2020deep} as the uncertified base model, trained on Gaussian noise with $\sigma=0.25$ as detailed in App.~\ref{subsection:training}. We use different parameters for the threshold function $T_{\mathrm{thresh}}$ (Eq \ref{eq:fthresh}) for \textsc{AdaptiveCertify} per dataset as described in App.~\ref{subsection:thresh}, which we found via a grid search that maximizes the $\cig$ metric. The evaluation metrics include $\cig$, $c\cig$, $\%\oslash$ (percentage of abstain pixels), or its complement: $\%\text{certified}$ (percentage of certified pixels), and $c\%\oslash$ (class-averaged $\%\oslash$). All details on the experimental setup can be found in App.~\ref{section:setup}.

\begin{figure*}[!ht]
\captionsetup[subfigure]{justification=centering}
\centering
\subfloat[]  
{\resizebox{0.33\textwidth}{!}{\includegraphics{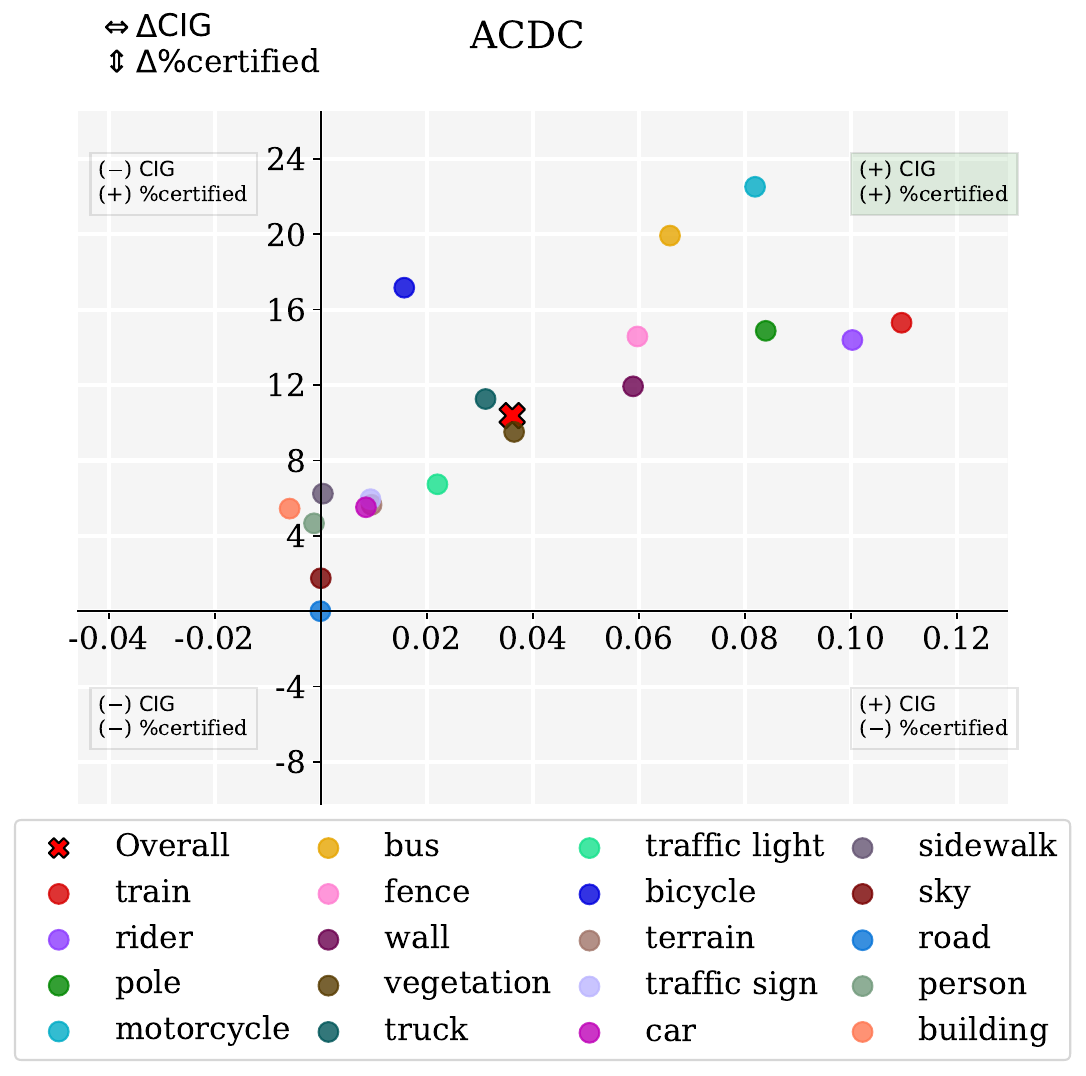}}}
\subfloat[]
{\resizebox{0.29\textwidth}{!}{\includegraphics{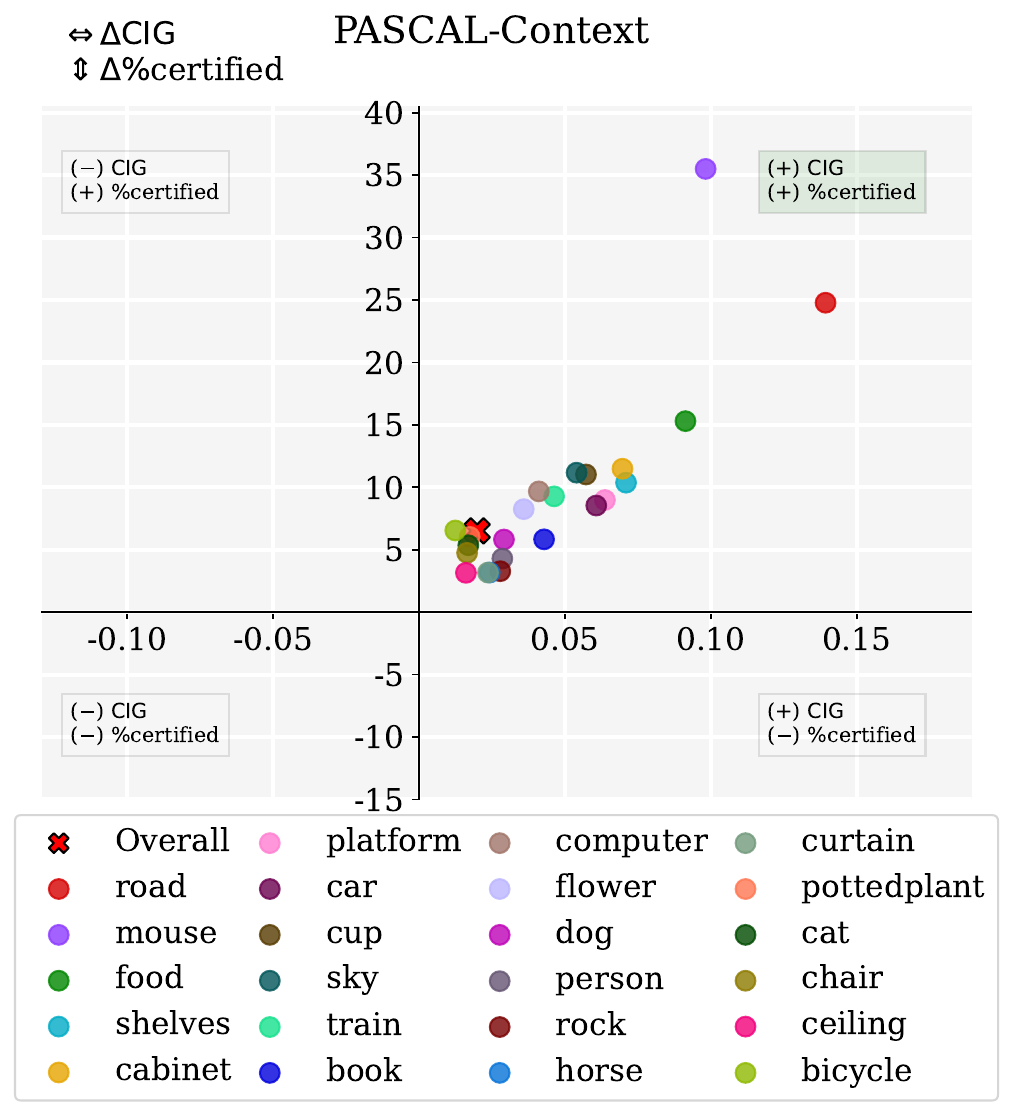}}}  
\subfloat[]  
{\resizebox{0.32\textwidth}{!}{\includegraphics{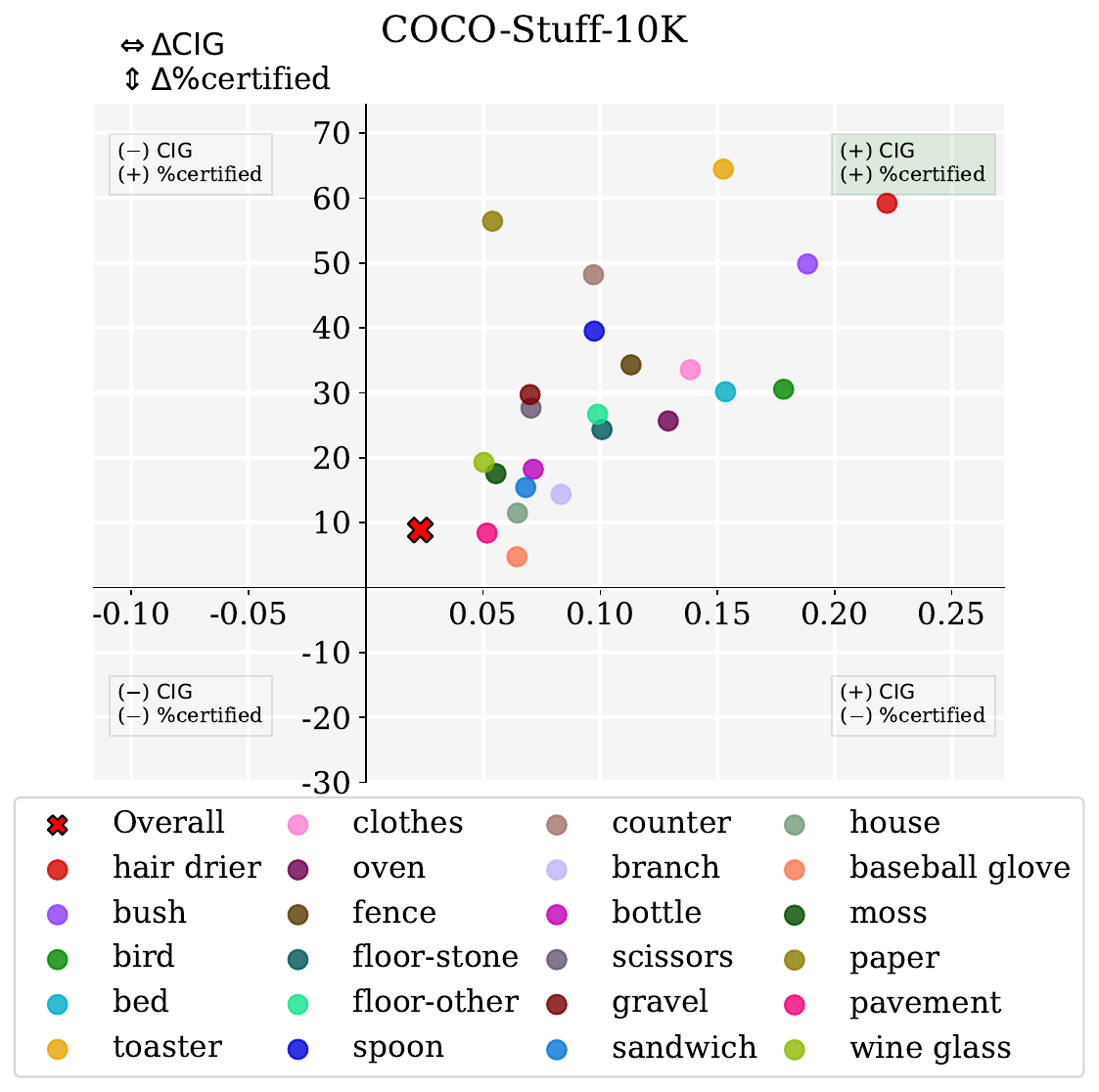}}} 
\caption{The performance of \textsc{AdaptiveCertify} against the baseline in terms of the difference in $\cig$ ($\Delta \cig$) and the certification rate ($\Delta \%$certified) across the 3 datasets w.r.t their top classes. ``Overall" indicates the class-average performance. Extensions of this figure to Cityscapes and all classes in PASCAL-Context and COCO-Stuff-10K datasets are in App. Figures \ref{fig:delta-cs}, \ref{fig:pascal-delta-ext} and \ref{fig:coco-delta-ext} under App.~\ref{subsection:delta}.}
\label{fig:delta-all}
\end{figure*}

We first investigate the overall performance of \textsc{AdaptiveCertify} against the baseline \textsc{SegCertify} across noise levels $\sigma$ and number of samples $n$ in Table \ref{tab:table}. On a high level, \textsc{AdaptiveCertify} consistently has a higher $\cig$ and lower $\%\oslash$ than \textsc{SegCertify}. Although increasing the noise level $\sigma$ degrades the performance in both algorithms, \textsc{AdaptiveCertify} abstains much less than \textsc{SegCertify}, while maintaining a higher $\cig$, at higher noise levels. The improvement in $\cig$ and $\%\oslash$ is highest on the COCO-Stuff-10K dataset, at $3.4\%$ and $35\%$. COCO-Stuff-10K has a large number of classes --171-- best highlighting the efficacy of our hierarchical certification approach. We investigate the overall performance in more detail, including the mIoU, $c\cig$ and $c\%\oslash$ metrics per dataset in App.~\ref{subsection:overall-performance}. %

We investigate the per-class performance of \textsc{AdaptiveCertify} and \textsc{SegCertify} in Figure \ref{fig:delta-all} in terms of $\cig$ and \%certified on the datasets: ACDC, PASCAL-Context and COCO-Stuff-10K. The average overall difference in performance shows that we certify ~$12\%$, $5\%$ and $10\%$ more of the pixels in all 3 datasets respectively, while achieving a higher $\cig$ by $\approx 0.04$ on ACDC and $\approx 0.02$ on the rest. Almost all of the classes lie in the quadrant where \textsc{AdaptiveCertify} outperforms \textsc{SegCertify} across both metrics (upper right quadrant), reaching a maximum improvement in $\cig$ of $+0.23$ in the class \class{hair drier} in COCO-Stuff-10K and $\%$certified increase of $+65$ percentage points in the class \class{toaser} in the same dataset. The performance remains the same (with a $\Delta$ of $0$) for leaf classes with no parent vertices at coarser hierarchy levels, such as \class{road} in ACDC (hierarchy in Figure \ref{fig:cs-h}), since by definition \textsc{AdaptiveCertify} is reduced to \textsc{SegCertify} in a single-level hierarchy.

\begin{figure}[!ht]
\resizebox{\columnwidth}{!}{\includegraphics[]{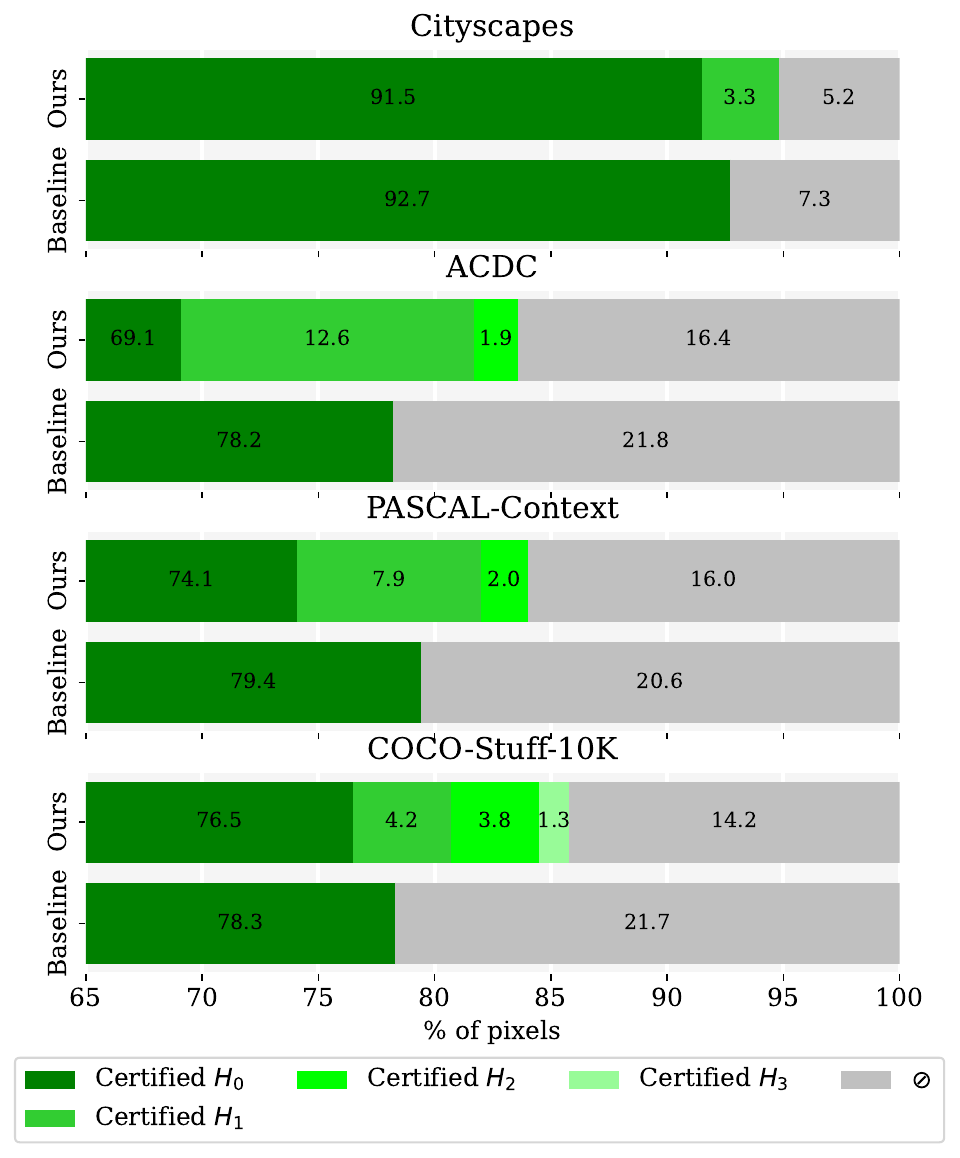}} 
\caption{The performance of our algorithm against the baseline in terms of percentage of abstain and certified pixels under different hierarchy levels. \textsc{SegCertify}, by definition, only uses $H_0$.}
\label{fig:dist_bar_all}
\end{figure}

To illustrate the hierarchical nature of \textsc{AdaptiveCertify}, we plot the pixel distribution across hierarchy levels in Figure \ref{fig:dist_bar_all}. We observe that both methods certify a comparable number of pixels at the finest level $H_0$. However, due the hierarchical structure of \textsc{AdaptiveCertify}, there is a notable advantage in certifying additional percentages (ranging from 2\% to 7\%) of pixels at higher hierarchy levels in the 4 datasets, where \textsc{SegCertify} opts to abstain. This effect is the strongest in the more challenging datasets ACDC, PASCAL-Context and COCO-Stuff-10K. Due to more fluctuating components, more pixels are assigned to coarser hierarchy levels.

Boundary pixels constitute a challenge in segmentation, prompting an analysis of the certification performance of \textsc{AdaptiveCertify} against a non-adaptive baseline (\textsc{SegCertify}) on them, as shown in Table \ref{tab:boundary-sub}. We explain the setup to isolate boundary pixels in App.~\ref{subsection:boundary}. Overall, \textsc{AdaptiveCertify} maintains a positive percentage improvement ($\%\Delta$) on both boundary and non-boundary pixels across all metrics. A higher percentage of boundary pixels is abstained from by both methods compared to the non-boundary pixels, with a maximum of $35\%$ in the challenging ACDC dataset by the baseline. Similarly, the $\cig$ of the boundary pixels is lower in both methods. This is attributed to the difficulty of segmenting boundary pixels as they mark label transitions in the segmentation map. However, the $\%\Delta$ improvement of \textsc{AdaptiveCertify} over the baseline in boundary pixels is on average higher than that of non-boundary pixels, with the exception of the PASCAL-Context dataset. This shows the effectiveness of the hierarchical grouping of labels in our adaptive method, especially on pixels the model is not confident about at object boundaries. We show qualitative analysis of a visual example in App.~\ref{subsection:boundary}.

\begin{table*}
\setlength\tabcolsep{0pt}
\centering
\begin{tabularx}{\textwidth}{@{}>{\raggedright\arraybackslash}X *{14}{>{\centering\arraybackslash}X}@{}}
\toprule
\multicolumn{3}{c}{} & \multicolumn{3}{c}{\textbf{Cityscapes}} & \multicolumn{3}{c}{\textbf{ACDC}} & \multicolumn{3}{c}{\textbf{PASCAL-Context}} & \multicolumn{3}{c}{\textbf{COCO-Stuff-10K}}\\
\midrule
 & &  & Baseline &  Ours &  $\%\Delta$ &  Baseline &  Ours &  $\%\Delta$ &  Baseline &  Ours &  $\%\Delta$ &  Baseline &  Ours &  $\%\Delta$ \\
\midrule
$\oslash\%\downarrow$ & & & $7.3$ & $5.2$  & {\fontsize{8}{0}\selectfont\textcolor{ForestGreen}{$28.8\%$}} & $21.8$ & $16.4$ & {\fontsize{8}{0}\selectfont\textcolor{ForestGreen}{$24.8\%$}} & $20.6$ & $16.1$ & {\fontsize{8}{0}\selectfont\textcolor{ForestGreen}{$21.8\%$}} & $21.7$ & $14.2$ & {\fontsize{8}{0}\selectfont\textcolor{ForestGreen}{$34.6\%$}} \\
$\oslash\%\text{boundary} \downarrow$  & & & $19.3$ & $13.7$  & {\fontsize{8}{0}\selectfont\textcolor{ForestGreen}{$29.0\%$}} & $35.0$ & $25.0$ & {\fontsize{8}{0}\selectfont\textcolor{ForestGreen}{$28.6\%$}} & $26.8$ & $22.2$ & {\fontsize{8}{0}\selectfont\textcolor{ForestGreen}{$17.2\%$}} & $26.4$ & $19.3$ & {\fontsize{8}{0}\selectfont\textcolor{ForestGreen}{$26.9\%$}} \\
$\oslash\%\text{non-boundary} \downarrow$  & & & $5.1$ & $3.7$  & {\fontsize{8}{0}\selectfont\textcolor{ForestGreen}{$27.5\%$}} & $20.2$ & $15.4$ & {\fontsize{8}{0}\selectfont\textcolor{ForestGreen}{$23.8\%$}} & $20.0$ & $15.5$ & {\fontsize{8}{0}\selectfont\textcolor{ForestGreen}{$22.5\%$}} & $20.5$ & $12.9$ & {\fontsize{8}{0}\selectfont\textcolor{ForestGreen}{$37.1\%$}} \\
$\cig\uparrow$  & &  & $0.89$ & $0.90$  & {\fontsize{8}{0}\selectfont\textcolor{ForestGreen}{$1.12\%$}} & $0.62$ & $0.63$ & {\fontsize{8}{0}\selectfont\textcolor{ForestGreen}{$1.61\%$}} & $0.55$ & $0.56$ & {\fontsize{8}{0}\selectfont\textcolor{ForestGreen}{$1.82\%$}} & $0.52$ & $0.54$ & {\fontsize{8}{0}\selectfont\textcolor{ForestGreen}{$3.85\%$}} \\
$\cig\text{-boundary}\uparrow$  & & & $0.71$ & $0.72$  & {\fontsize{8}{0}\selectfont\textcolor{ForestGreen}{$1.41\%$}} & $0.43$ & $0.45$ & {\fontsize{8}{0}\selectfont\textcolor{ForestGreen}{$4.65\%$}} & $0.37$ & $0.38$ & {\fontsize{8}{0}\selectfont\textcolor{ForestGreen}{$2.70\%$}} & $0.44$ & $0.46$ & {\fontsize{8}{0}\selectfont\textcolor{ForestGreen}{$4.55\%$}} \\
$\cig\text{-non-boundary}\uparrow$ & & & $0.92$ & $0.93$  & {\fontsize{8}{0}\selectfont\textcolor{ForestGreen}{$1.09\%$}} & $0.64$ & $0.65$ & {\fontsize{8}{0}\selectfont\textcolor{ForestGreen}{$1.56\%$}} & $0.56$ & $0.58$ & {\fontsize{8}{0}\selectfont\textcolor{ForestGreen}{$3.57\%$}} & $0.54$ & $0.56$ & {\fontsize{8}{0}\selectfont\textcolor{ForestGreen}{$3.70\%$}} \\
\bottomrule
\end{tabularx}
\caption{Mean per-pixel certification performance of \textsc{AdaptiveCertify} against \textsc{SegCertify} over the first 100 images from each dataset. $\oslash\%$boundary$=\frac{\text{number of $\oslash$ and boundary pixels}}{\text{number of boundary pixels}}$ and $\cig\text{-boundary}=\cig\text{ of boundary pixels only}$, and a similar logic follows for non-boundary metrics.}
\label{tab:boundary-sub}

\end{table*}

\begin{figure}[!ht]
{\resizebox{\columnwidth}{!}{\includegraphics{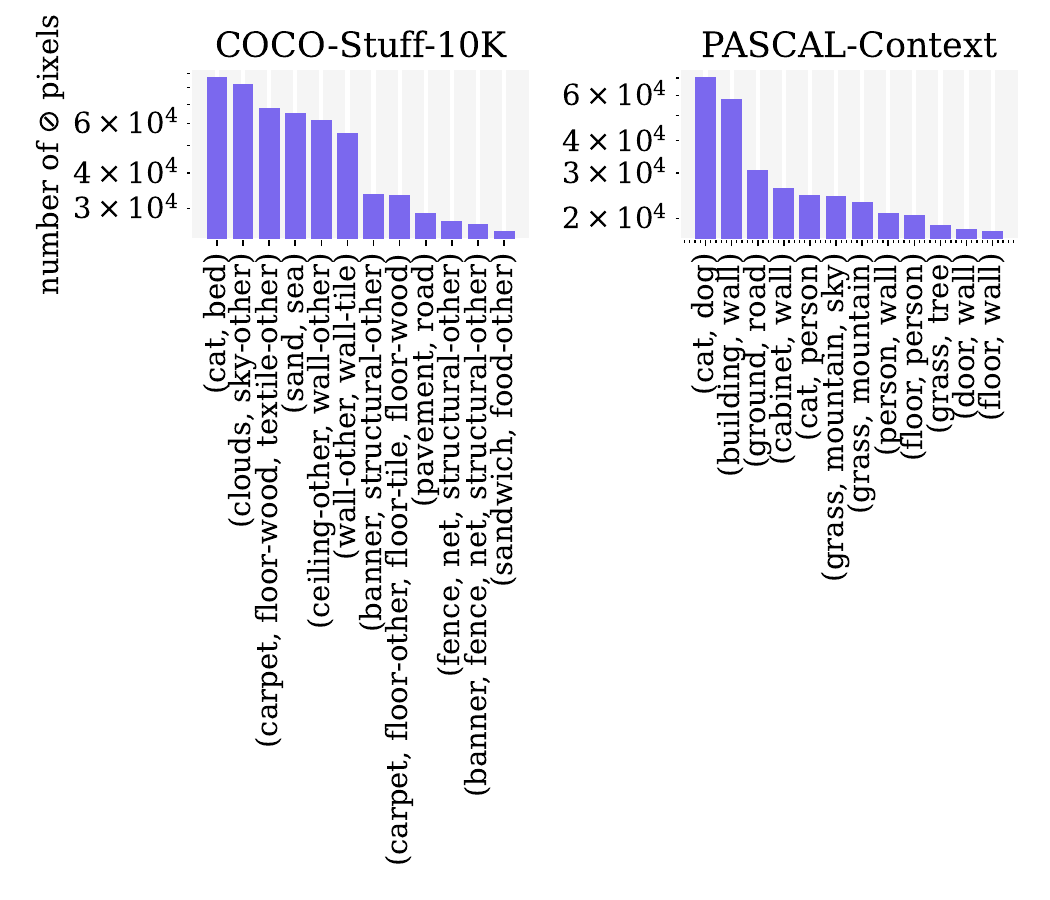}}}
\caption{The frequency of the top-most sets of classes the model fluctuates between in abstain pixels by the baseline on the COCO-Stuff-10K and PASCAL-Context datasets.}
\label{fig:fluct-sub}
\end{figure}

Abstentions occur due to model uncertainty caused by noise perturbations, indicating a lack of confidence in a single top class. We investigate the fluctuating output classes of the model by visualizing recurring class sets in unstable components (e.g., abstain pixels by \textsc{SegCertify}) in two datasets, shown in Figure \ref{fig:fluct-sub}. We observe these recurring sets aligning with higher-level semantic concepts; for instance, (\class{grass}, \class{mountain}, \class{sky}) are grouped under \class{nature} in PASCAL-Context hierarchy in App. Figure \ref{fig:pascal-h}, while (\class{dirt}, \class{grass}, \class{gravel}, \class{ground}-\class{other}, \class{sand}) fall under \class{ground} in COCO-Stuff \citep{cocostuff}. Our approach groups fluctuating classes under broader labels in a semantic hierarchy, leading to fewer abstentions.

\section{Conclusion}
In this paper, we investigate the problem of high abstain rates in common certification techniques that rely on a flat hierarchy. Based on that, we introduce adaptive hierarchical certification for semantic segmentation. We mathematically formulate the problem and propose \textsc{AdaptiveCertify}, that solves three main challenges: finding unstable components, adaptive sampling from multiple hierarchy levels, and evaluating the results using the Certified Information Gain metric. Instead of abstaining for unstable components, \textsc{AdaptiveCertify} relaxes the certification to a coarser hierarchy level. It guarantees an abstain rate less than or equal to non-adaptive versions while maintaining a higher $\cig$ across different noise levels and number of samples. Considering that boundary pixels constitute a challenge in segmentation, \textsc{AdaptiveCertify}'s improvement is widespread across both boundary and non-boundary pixels. The formulation of our hierarchical certification method is general and can adopt any hierarchy graph.

\clearpage
\section*{Impact Statement}

In this paper, we introduced adaptive hierarchical certification for semantic segmentation, in which certification can be within a multi-level hierarchical label space, which lowers the abstain rate and increases the certified information gain compared to conventional methods. We adapt and relax the certification for challenging pixels, by certifying them to a coarser label in a semantic hierarchy, rather than abstaining. Our adaptive hierarchical certification method not only addresses the limitations of conventional certification methods but also presents ethical considerations and potential societal consequences in domains crucial to human well-being. The overall impact is assessed to be  positive as it contributes to understanding, methodology, and mitigation of robustness issues with current AI/ML methods.

\myparagraph{Societal impact:} Image semantic segmentation is of importance to safety-critical applications, including autonomous driving, medical imaging, and video surveillance. Through hierarchical certification of segmentation models deployed in these domains, we provide more certified information gain on average, particularly in challenging images. This is particularly noteworthy as conventional certification methods often encounter significant abstain rates in such complex scenarios. For instance, in the context of autonomous driving, it is more advantageous for the system to ascertain whether a group of pixels is drivable or not, rather than refraining from certification due to the constraints of attempting to certify them within numerous fine-grained classes. Collaboration with stakeholders in autonomous driving, medical research, and surveillance technology can facilitate the integration of the idea of hierarchical certification, leading to enhanced decision support systems and contributing to the overall safety and efficiency of these applications.

\myparagraph{Ethical considerations:} Our adaptive hierarchical certification provides more certified semantically meaningful information in image semantic segmentation, which contributes to ethical decision-making processes in the aforementioned safety-critical applications that rely on semantic segmentation.

\myparagraph{Conclusion} In conclusion, our paper not only introduces a novel approach to adaptive hierarchical certification for image semantic segmentation but also emphasizes its critical role in safety-critical applications. By addressing ethical considerations and anticipating future societal consequences, our work serves as a catalyst for advancements in technology. Overall, we foresee a positive impact, as the contribution addresses short comings of AI/ML method w.r.t. robustness and reliability that will ultimately lead to safer AI/ML systems.

\section*{Acknowledgements}
We very much appreciate the diligent and constructive reviews by all reviewers, and believe the additional insights gained in preparing the rebuttal, and expanding the analysis in our work, significantly strengthen our paper. 

This work was partially funded by ELSA – European Lighthouse on Secure and Safe AI funded by the European Union under grant agreement No. 101070617. Views and opinions expressed are however those of the authors only and do not necessarily reflect those of the European Union or European Commission. Neither the European Union nor the European Commission can be held responsible for them.

\bibliography{references}
\bibliographystyle{icml2024}

\newpage
\appendix
\onecolumn

\section{Algorithm functions}\label{subsection:algos}
The method \textsc{SamplePosteriors} retrieves $n_0$ posteriors per component: ${Ps}^0_1, \ldots, {Ps}^0_N$, such that ${Ps}^0_i$ is a set of $n_0$ posterior vectors $\in [0, 1]^{|\mathcal{Y}|}$ for the $i^\text{th}$ component, as outline in Algorithm \ref{alg:sample-posteriors}. The function $f_{i, \mathrm{seg}}$ returns the posteriors $Ps_i$ of the $i^\text{th}$ component, such that $f_{\mathrm{seg}}$ is the segmentation head of $f$.
\begin{algorithm}[htbp]
\caption{\textsc{SamplePosteriors}: algorithm to sample posteriors given noise $\sigma$}\label{alg:sample-posteriors}
\begin{algorithmic}
\FUNCTION{\textsc{SamplePosteriors}($f$, $x$, $n$, $\sigma$)}
 \STATE $f_{\mathrm{seg}} \gets$ \text{get segmentation head from } $f$
      \FOR{$j \gets {1, \ldots, n}$}
      \FOR{$i \gets {1, \ldots, N}$}
        \STATE \text{draw random noise $\epsilon \sim \mathcal{N}(0, \sigma^2 I)$}
        \STATE $Ps_i^j \gets f_{i,{\mathrm{seg}}}(x+\epsilon)$
        \ENDFOR
      \ENDFOR
\STATE \textbf{return} ($Ps^n_1, \ldots , Ps^n_N$)
\ENDFUNCTION
\end{algorithmic}
\end{algorithm}

We apply multiple hypothesis testing, similar to \citet{fischer}, following the Bonferroni method \cite{bonferroni1936teoria} to reject (certify) or accept (abstain by overwriting $\hat{v}_i$ with $\oslash$) the null hypotheses of components while maintaining an overall type I error probability of $\alpha$. The \textsc{HypotheseTesting} function as described is outlined in Algorithm \ref{alg:h-testing}.
\begin{algorithm}[htbp]
\caption{\textsc{HypothesesTesting}: algorithm to perform multiple hypotheses testing while bounding the Type I error rate by $\alpha$ following the Bonferroni method \citep{bonferroni1936teoria}}\label{alg:h-testing}
\begin{algorithmic}
\FUNCTION{\textsc{HypothesesTesting}($\alpha$, $\oslash$, ($pv_1$,\ldots ,$pv_N$), ($\hat{v}_1, \ldots, \hat{v}_N$))}
    \FOR{$i \gets {1, \ldots, N}$}
    \IF{$pv_i > \frac{\alpha}{N}$}
    \STATE $\hat{v}_i \gets \oslash$
    \ENDIF
    \ENDFOR
\STATE \textbf{return} $\hat{v}_1, \ldots, \hat{v}_N$
\ENDFUNCTION
\end{algorithmic}
\end{algorithm}

\section{Experimental setup}\label{section:setup}
We outline the experimental setup for our evaluation experiments that are presented in Section \ref{sec:experiments}. We first discuss the datasets used and their corresponding semantic DAG hierarchies (App.~\ref{subsection:datasets}). Note that the inference is invoked on images with their original dimension without scaling. Next, we describe the training setup for the models used per dataset (App.~\ref{subsection:training}). Lastly, we list the values for all relevant evaluation parameters, such as the threshold function parameters (App.~\ref{subsection:thresh}) and the certification settings (App.~\ref{subsection:cert-params}).

\subsection{Datasets and Hierarchies}\label{subsection:datasets}
\paragraph{Cityscapes} Cityscapes \cite{cityscapes} is a large-scale scene-understanding dataset of urban scene images ($1024\times 2048$ px) across 50 different cities in Germany and surrounding regions. The dataset provides pixel-level annotations for 30 classes. Our evaluation focuses on the official mode of semantic segmentation benchmarks with 19 common classes found in urban street scenes. The hierarchy DAG on top of these 19 classes is illustrated in Figure \ref{fig:cs-h} and is used in all \textsc{AdaptiveCertify} experiments related to this dataset.

\begin{figure}[!ht]
\centering
\resizebox{0.7\textwidth}{!}{\includegraphics[]{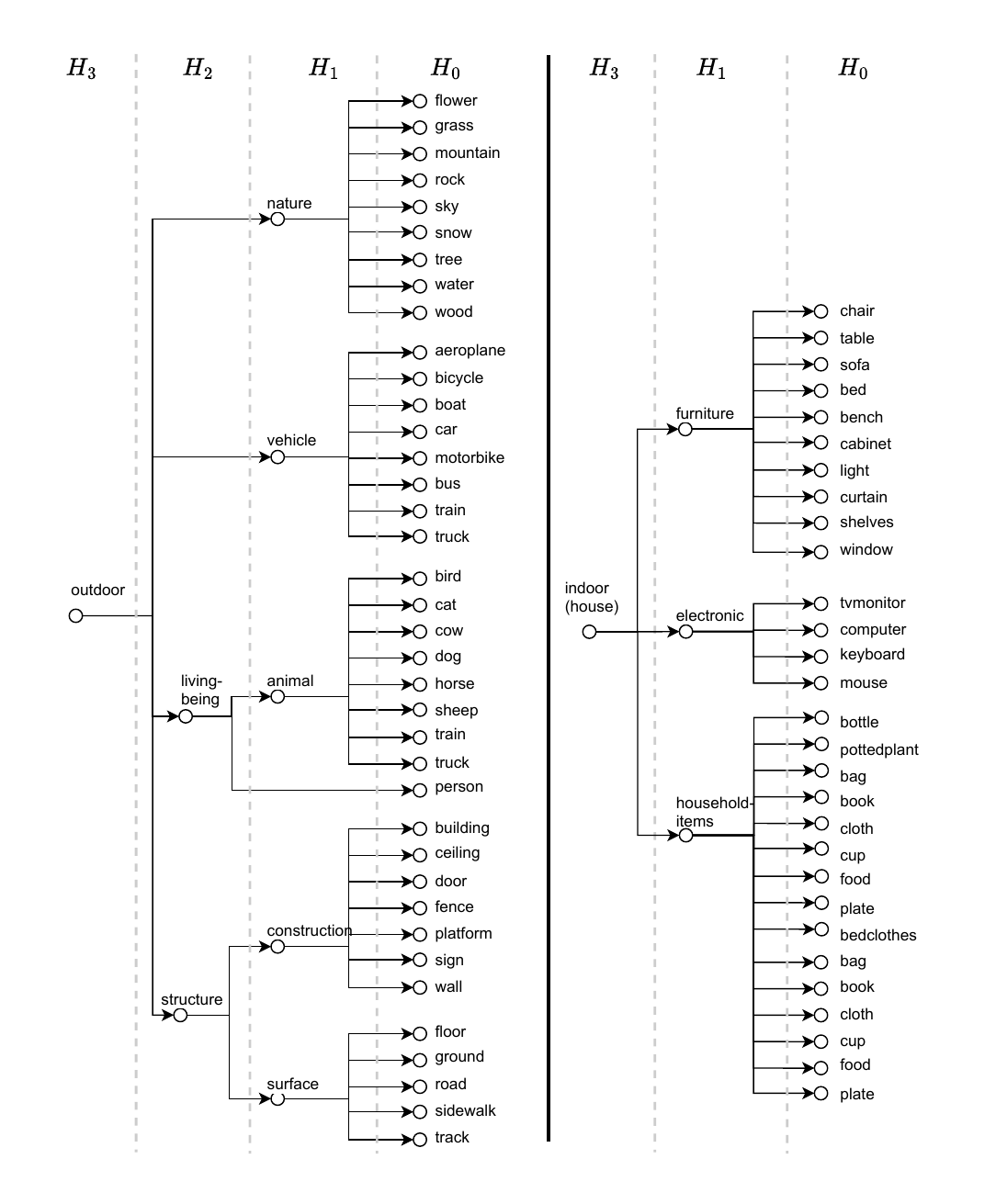}}  
\caption{A DAG graph representing a semantic hierarchy on top of the $59$ foreground classes of PASCAL-Context \citep{pascal}}
\label{fig:pascal-h}
\end{figure}

\paragraph{PASCAL-Context} PASCAL-Context is a scene-understanding dataset with detailed pixel-wise semantic labels, an extension of the PASCAL VOC 2010 dataset encompassing over 400 classes \cite{pascal}. We evaluate this dataset with 59 foreground classes in a setup similar to the prior work  from \citet{fischer}. The semantic hierarchy designed for the 59 classes is illustrated in Figure \ref{fig:pascal-h} for use in our \textsc{AdaptiveCertify} experiments.

\paragraph{Adverse Conditions Dataset with Correspondences (ACDC)} ACDC is a challenging traffic scene dataset featuring images captured under four adverse visual conditions: snow, rain, fog, and nighttime \cite{acdc}. It includes 19 classes identical to the evaluation classes in Cityscapes ($H_0$ in Appendix Figure \ref{fig:cs-h}), and we adopt the Cityscapes hierarchy DAG for this dataset.

\paragraph{Common Objects in COntext-Stuff (COCO-Stuff)} COCO-Stuff \citep{cocostuff} is a scene understanding dataset, that is an extension of the COCO dataset \cite{coco}, which addresses the segmentation of pixels as either thing or stuff classes. It has 172 categories (80 things, 91 stuff and 1 unlabelled). Our evaluation utilizes the common mode with 171 categories, excluding the unlabelled class. We work with the COCO-Stuff-10K v1.1 subset, consisting of 9k training and 1k validation splits. The pre-defined hierarchy for things and stuff officially provided by the dataset \cite{cocostuff} is used for the DAG hierarchy.

\subsection{Models and Training}\label{subsection:training}

For the Cityscapes, ACDC and PASCAL-Context, HrNetV2 \citep{wang2020deep, hrnetocr} is used, with the HRNetV2-W48 backbone. We use the weights provided by \citep{fischer} in their official paper PyTorch \cite{paszke2019pytorch} implementation, which is the result of training the model on a Gaussian noise of $\sigma=0.25$ following a similar training procedure to that of the PyTorch implementation of HrNetV2. It's important to note that for Cityscapes and ACDC, we employ the same model trained solely on Cityscapes data. This choice is intentional to evaluate the adaptive hierarchical certification method under slight domain shifts present in ACDC. The clean accuracy of the final HrNetV2 model is $90\%$, $61\%$, and $58\%$ on the three datasets, respectively.

\begin{figure*}[!ht]
\captionsetup[subfigure]{justification=centering}
\centering
\subfloat[]
{\resizebox{0.45\textwidth}{!}{\includegraphics{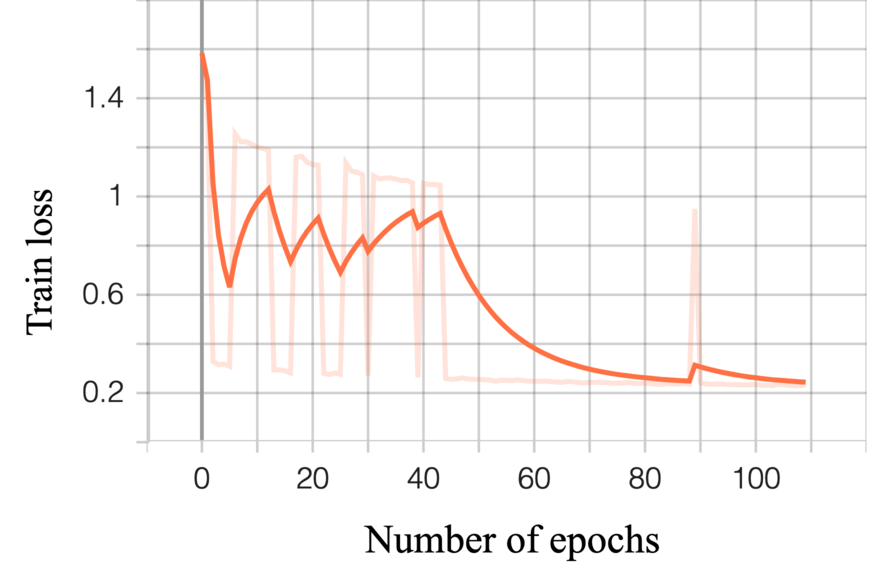}}}  
\hspace{0.5em}
\subfloat[]  
{\resizebox{0.45\textwidth}{!}{\includegraphics{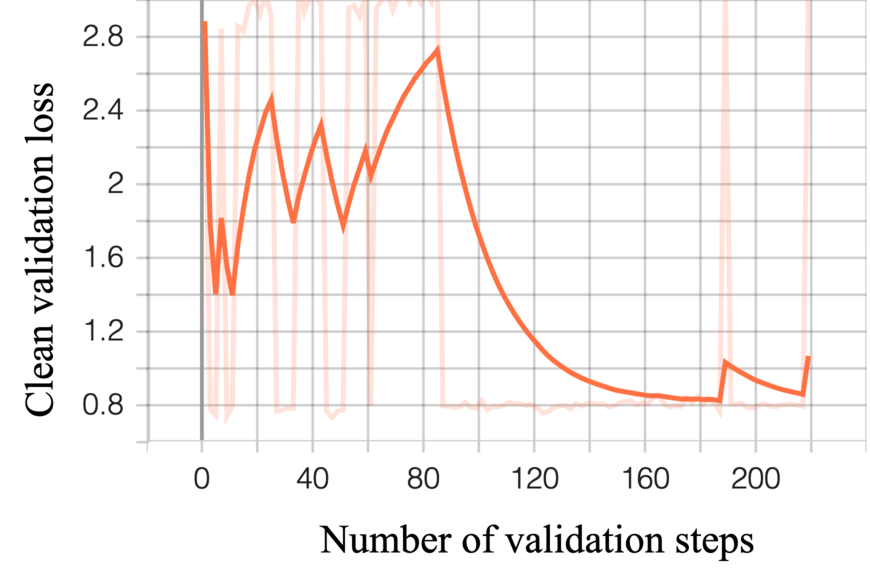}}} 
\hspace{0.5em}
\\
\subfloat[]  
{\resizebox{0.5\textwidth}{!}{\includegraphics{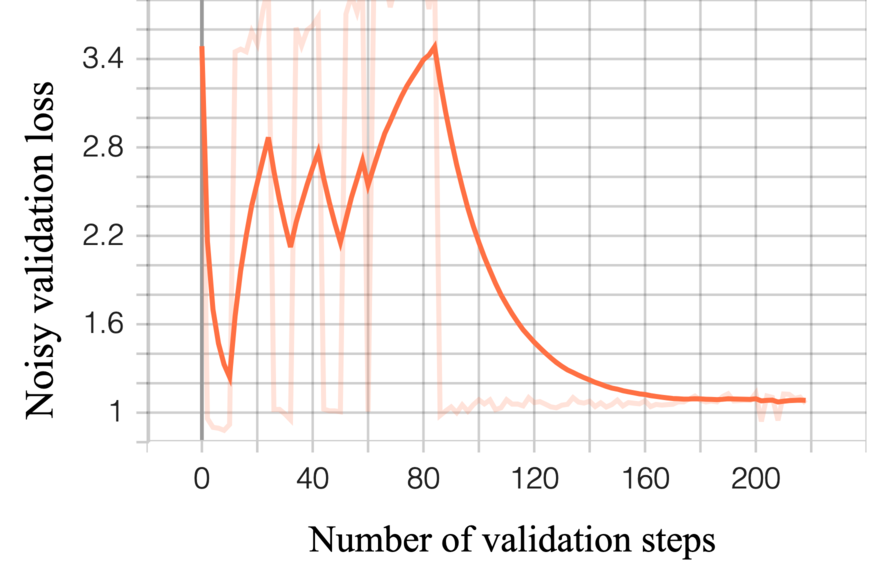}}}
\caption{The train and validation loss during the training of HrNetV2 model \cite{wang2020deep, hrnetocr} on the COCO-Stuff-10K dataset \citep{cocostuff} on noise with $\sigma=0.25$. \resolved{\bernt{fonts in the is quite small really - would be better to have larger font or increase the size of the plots themselves}\alaa{I increased the size of the graphs}}}
\label{fig:coco-loss}
\end{figure*}

\begin{figure*}[!ht]
\captionsetup[subfigure]{justification=centering}
\centering
\subfloat[]
{\resizebox{0.42\textwidth}{!}{\includegraphics{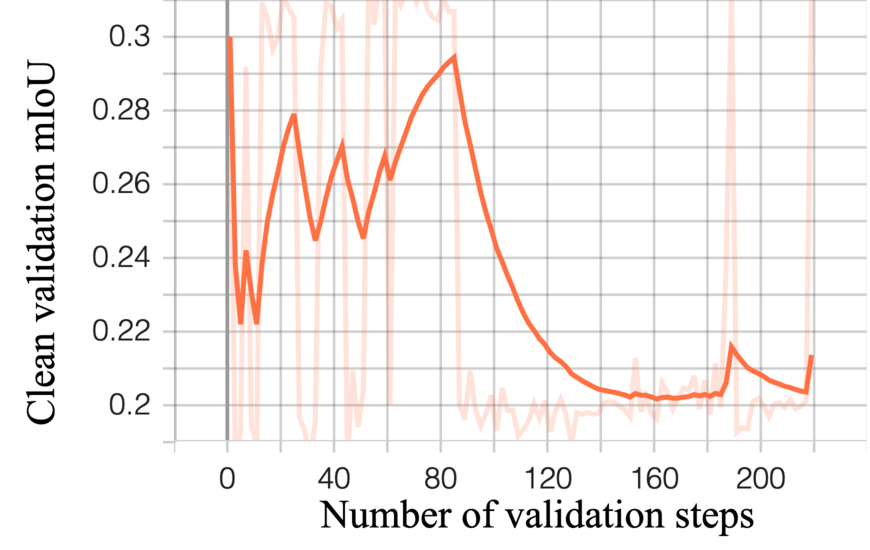}}}  
\hspace{0.1em}
\subfloat[]  
{\resizebox{0.42\textwidth}{!}{\includegraphics{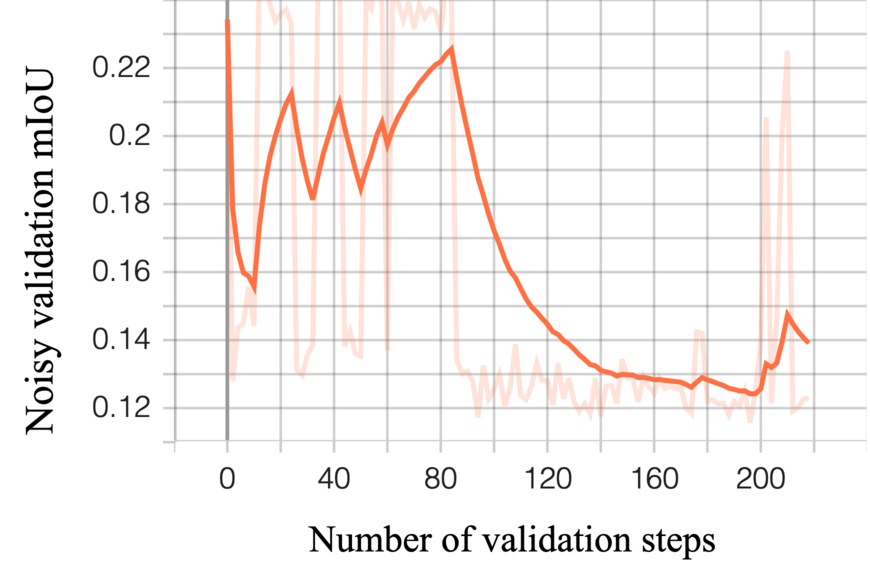}}} 
\caption{The clean and noisy validation mIoU during the training of the HrNetV2 model on the COCO-Stuff-10K dataset.}
\label{fig:coco-miou}
\end{figure*}

\begin{figure*}[!ht]
\captionsetup[subfigure]{justification=centering}
\centering
\subfloat[]
{\resizebox{0.42\textwidth}{!}{\includegraphics{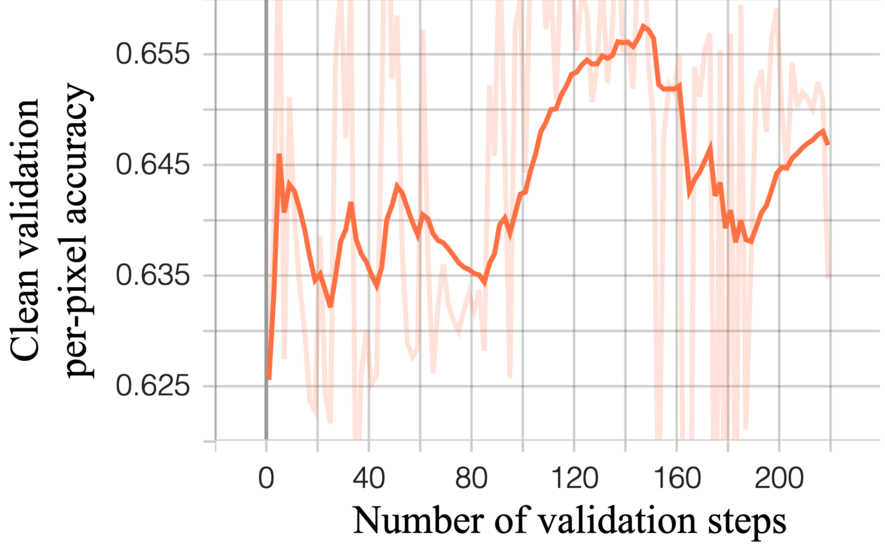}}}  
\hspace{0.1em}
\subfloat[]  
{\resizebox{0.42\textwidth}{!}{\includegraphics{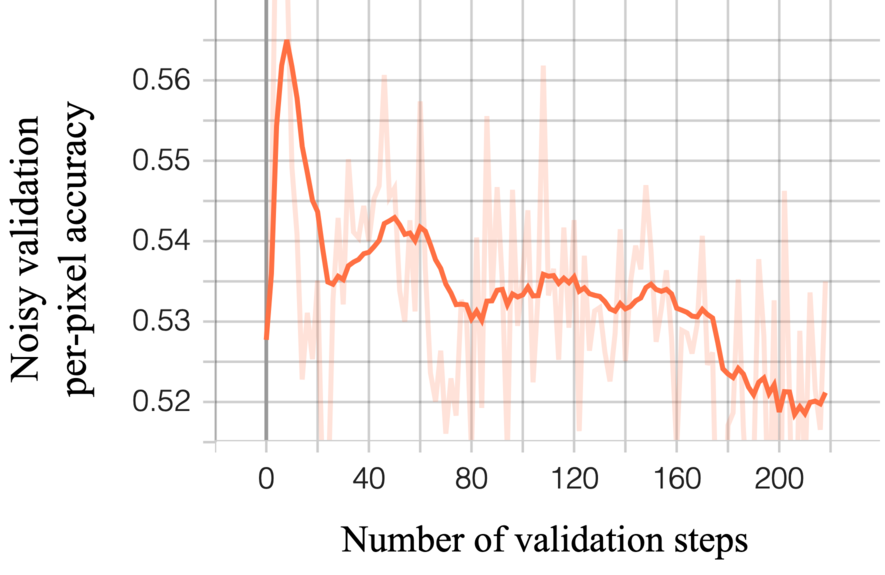}}} 
\caption{The clean and noisy validation mean per-pixel accuracy during the training of the HrNetV2 model on the COCO-Stuff-10K dataset.}
\label{fig:coco-acc}
\end{figure*}

For the COCO-Stuff-10K dataset \citep{cocostuff}, we used the HrNetV2 model \cite{wang2020deep, hrnetocr} with the HrNetV2-W48 Paddle Cls pre-trained backbone that follows the Object-Contextual Representations (OCR) approach, which is available in the official PyTorch implementation of the HrNetV2 paper \cite{paszke2019pytorch}. We follow the same outlined training procedure, except by adding a Gaussian noise of $\sigma=0.25$ in an alternating manner across the batches for the same number of $110$ epochs. The final model performance on the clean split has a mean per-pixel accuracy of 62.77\% and an mIoU of 0.3146. Meanwhile, on the noisy validation split, the mean per-pixel accuracy is 53.43\% and the mIoU is 0.2436. We validate twice every epoch on both the clean and noisy ($\sigma=0.25$) validation splits. During validation, we calculate the following metrics: the mean per-pixel accuracy, the mean accuracy, and the mean intersection over union (mIoU). The batch sizes used for training and validation were $12$ and $1$ respectively, per GPU. We validate on non-resized images with a scale of $1$, and also show the certification results on them. In Figure \ref{fig:coco-loss}, the training loss is shown in subfigure (a), with validation loss shown for both the clean and noisy validation splits in subfigures (b) and (c). Additionally, Figures \ref{fig:coco-miou} and \ref{fig:coco-acc} illustrate validation mIoU and mean per-pixel accuracy on clean and noisy validation splits during training.

\subsection{Threshold Function}\label{subsection:thresh}

We conducted a grid search to find the best threshold function $T_{\text{thresh}}$ parameters to use per dataset. The best parameters are chosen such that they score the maximum $\cig$ on the first 100 samples of the test set of the dataset, fixing the rest of the certification parameters to $n_0=10$, $\tau=0.75$, $\sigma=0.25$ and $\alpha=0.001$.

In the following Table \ref{tab:thresh}, we show the threshold functions used by \textsc{AdaptiveCertify} for all four datasets.

\begin{table}[!ht]
\caption{The threshold function $T_{\text{thresh}}$ parameters used throughout our experiments in Section \ref{sec:experiments}.}
\begin{tabularx}{\textwidth}{@{} *{2}{X} @{}}
\toprule %
 & \textbf{Threshold function parameters} \\
\midrule %
Cityscapes & (0, 0,  0.25) \\
PASCAL-Context & (0, 0.1,  0.4) \\
ACDC & (0, 0.05, 0.3) \\
COCO-Stuff-10K & (0, 0.3,  0.7) \\
\bottomrule %
\end{tabularx}
\label{tab:thresh}
\end{table}

\begin{figure}[!ht]
\centering
{\resizebox{0.6\textwidth}{!}{\includegraphics{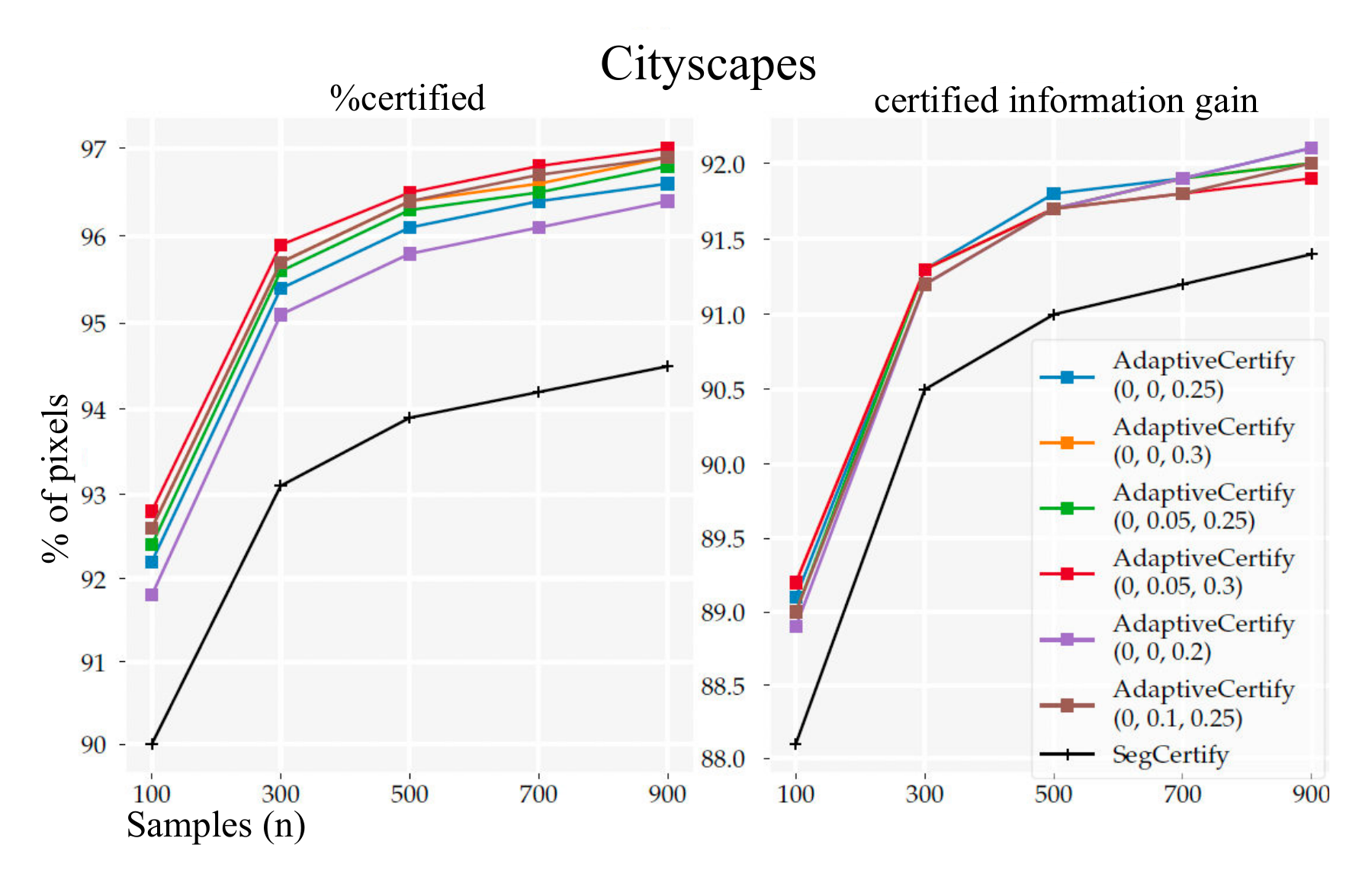}}} 
\caption{The performance of multiple versions of \textsc{AdaptiveCertify} by varying the threshold function parameters $T_\mathrm{thresh}$ against \textsc{SegCertify} in terms of CIG and \%certified on Cityscapes. The legend is in a descending order of the mean performance. Over $63$ threshold functions were used in the grid search, but we are plotting only some of them for clarity.}
\label{fig:threshold-search}
\end{figure}

We show an example of the grid search results on Cityscapes by showing the performance of different threshold functions across different number of samples $n$ in Figure \ref{fig:threshold-search}. We find that the best thresholds for Cityscapes that give the highest mean certified information gain compared to the rest are $(0, 0, 0.25)$ (as also mentioned in Table \ref{tab:thresh}).

\subsection{Certification parameters}\label{subsection:cert-params}
Unless stated otherwise, all certification results use the values $\sigma=0.25$, $\tau=0.75$ and $\alpha=0.001$ in both our algorithm \textsc{AdaptiveCertify} and the baseline \textsc{SegCertify}, and the metrics are the per-pixel mean over the first $100$ images in each dataset. 

\section{Results extended}
\subsection{Semantic fluctuations (Extended)}\label{subsection:fluctuations}

\begin{figure}[!ht]
\centering
\resizebox{\textwidth}{!}{\includegraphics[]{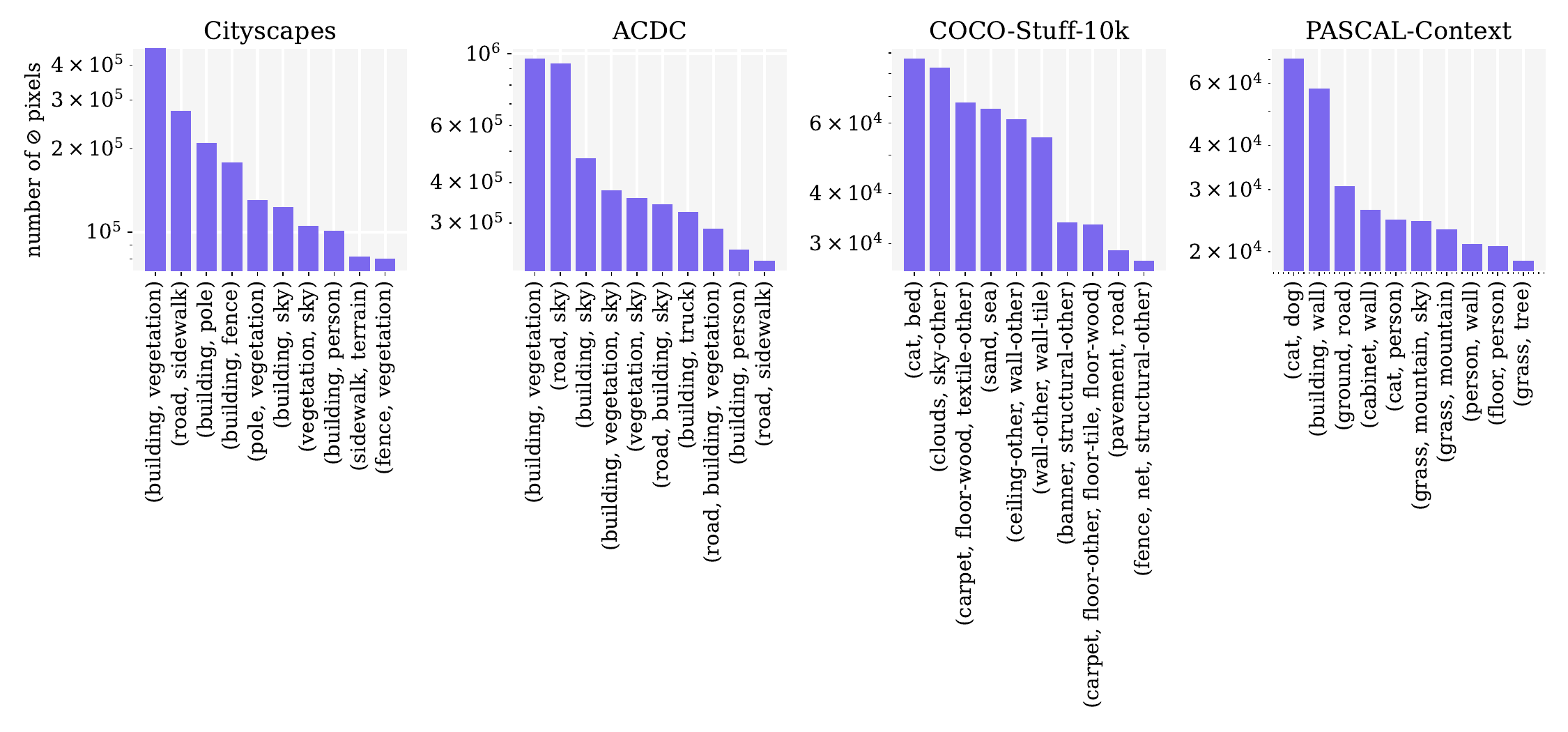}}  
\caption{The frequency of the sets of classes the model fluctuates between in abstain pixels across 4 datasets: Cityscapes, ACDC, COCO-Stuff-10K and PASCAL-Context. The y-axis is a log scale. This is the result of running on 40 images per dataset.}
\label{fig:fluct}
\end{figure}

The reason why randomized smoothing abstains from pixels is due to the fluctuating output of the model on perturbing such pixels with noise. It implies that the model is not confident about a single top class. Our method relies on grouping such fluctuating classes under coarser labels in higher levels in a semantic hierarchy, and thus abstaining less. In this section, we look into how those unstable components fluctuate amongst classes that are semantically related. We show the most recurring sets of classes in unstable components (e.g., abstain pixels by \textsc{SegCertify}) in all 4 datasets in Figure \ref{fig:fluct}. 

We observe that the topmost recurring sets of classes can be grouped under a higher level semantic concept. For example, we have as part of the top 10 sets:
\begin{itemize}
    \item (\class{clouds}, {sky-other}): grouped under \class{sky} in the the pre-defined COCO-Stuff hierarchy \cite{cocostuff}.
    \item (\class{dirt}, \class{grass}, \class{gravel}, \class{ground}-\class{other}, \class{sand}): grouped under \class{ground} in the predefined COCO-Stuff hierarchy \cite{cocostuff}.
    \item (\class{building}, \class{vegetation}): grouped under \class{Construction and Vegetation} in the Cityscapes hierarchy in Figure \ref{fig:cs-h}.
    \item (\class{building}, \class{person}): grouped under \class{obstacle} in the Cityscapes hierarchy in Figure \ref{fig:cs-h}.
    \item (\class{floor}, \class{ground}): grouped under \class{surface} in the PASCAL-Context hierarchy we propose and use in App. Figure \ref{fig:pascal-h}.
\end{itemize}

\subsection{Overall Performance of \textsc{AdaptiveCertify} (Extended)}\label{subsection:overall-performance}
We examine the overall performance of \textsc{AdaptiveCertify} in comparison to the current state-of-the-art \textsc{SegCertify} under varying noise levels ($\sigma$) and sample sizes ($n$). Results across all four datasets are presented in Tables \ref{tab:cityscapes-table}, \ref{tab:pascal_ctx-table}, \ref{tab:acdc-table}, and \ref{tab:cocostuff-table}. In essence, \textsc{AdaptiveCertify} consistently demonstrates higher Certified Information Gain and lower abstention rates compared to \textsc{SegCertify} across all datasets. Despite the performance degradation observed with increased noise levels ($\sigma$) for both algorithms, \textsc{AdaptiveCertify} notably exhibits reduced abstention rates while maintaining higher Certified Information Gain at even higher noise levels.

\paragraph{Increase in $\cig$} Across all four datasets, there is a minimum $\cig$ increase of $1.1\%$. Notably, the COCO-Stuff-10K dataset in Table \ref{tab:cocostuff-table} shows a maximum increase of $3.4\%$ under $\sigma=0.25$, $n=100$, and $\tau=0.75$. This substantial improvement in performance on the COCO-Stuff-10K dataset can be attributed to its large number of classes --171--, showcasing the hierarchy's grouping effect of many fine-grained classes under coarser labels. Similarly, the second-largest performance increase of $1.8\%$ is observed on the PASCAL-Context dataset, also attributed to it containing the second largest class set of 59 classes.

\paragraph{Decrease in the abstention rate} Across all four datasets, we observe a minimum decrease of $20\%$ in the abstention rate on the PASCAL-Context dataset (Table \ref{tab:pascal_ctx-table}), with a maximum decrease of $35\%$ on the COCO-Stuff-10K dataset (Table \ref{tab:cocostuff-table}). This significant decrease in the abstention rate can be attributed to the dataset's large number of classes, resulting in more unstable components from which the baseline tends to abstain. In contrast, our approach relaxes the certification to a coarser class, leading to a substantial improvement in performance. This improvement is further evidenced by the maximum increase in $\cig$, as discussed earlier.

\begin{table*}[!ht]
    \begin{tabularx}{\textwidth}{lXrrllllll}
    \toprule
    \multicolumn{5}{l}{} & \multicolumn{5}{c}{Cityscapes} \\
    \cmidrule{6-10}
    &  & \multicolumn{1}{l}{$\sigma$} & \multicolumn{1}{l}{$R$} &  & $\cig\uparrow$ & $c\cig\uparrow$  & $\%\oslash\downarrow$ &  $c\%\oslash\downarrow$ & mIoU $\uparrow$ \\
    \midrule
    & Uncertified HrNet & \multicolumn{1}{l}{-} & \multicolumn{1}{l}{-} &  & $0.90$ & $0.47$ & $-$ & $-$ & $0.39$ \\
    \midrule
    \multirow{7}{*}{\begin{tabular}[c]{@{}l@{}}$n = 100,$\\ $\tau=0.75$\end{tabular}} & \multirow{3}{*}{\textsc{SegCertify}} & $0.25$ & $0.17$ &  & $0.89$ & $0.65$ & $7$ & $21$ & $0.37$ \\
    &  & $0.33$ & $0.22$ &  & $0.81$ & $0.51$ & $14$ & $34$ & $0.30$ \\
    &  & 0.50 & 0.34 &  & $0.41$ & $0.11$ & $26$ & $35$ & $0.05$ \\
    & \multicolumn{8}{l}{} \\
    & \multirow{3}{*}{\textsc{AdaptiveCertify}} & 0.25 & 0.17 &  & \textbf{0.90 {\fontsize{6}{0}\selectfont\textcolor{ForestGreen}{$1.1\%$}}} & \textbf{0.66 {\fontsize{6}{0}\selectfont\textcolor{ForestGreen}{$1.5\%$}}} & \textbf{5 {\fontsize{6}{0}\selectfont\textcolor{ForestGreen}{$28.6\%$}}} & \textbf{16 {\fontsize{6}{0}\selectfont\textcolor{ForestGreen}{$23.8\%$}}} & $-$ \\
    &  & 0.33 & 0.22 &  & \textbf{0.83 {\fontsize{6}{0}\selectfont\textcolor{ForestGreen}{$2.5\%$}}} & \textbf{0.53 {\fontsize{6}{0}\selectfont\textcolor{ForestGreen}{$3.9\%$}}} & \textbf{10 {\fontsize{6}{0}\selectfont\textcolor{ForestGreen}{$28.6\%$}}} & \textbf{27 {\fontsize{6}{0}\selectfont\textcolor{ForestGreen}{$20.6\%$}}} & $-$ \\
    &  & 0.50 & 0.34 &  & \textbf{0.44 {\fontsize{6}{0}\selectfont\textcolor{ForestGreen}{$7.3\%$}}} & \textbf{0.13 {\fontsize{6}{0}\selectfont\textcolor{ForestGreen}{$18.2\%$}}} & \textbf{15 {\fontsize{6}{0}\selectfont\textcolor{ForestGreen}{$42.3\%$}}} & \textbf{20 {\fontsize{6}{0}\selectfont\textcolor{ForestGreen}{$42.9\%$}}} & $-$ \\
    \midrule
    \multirow{7}{*}{\begin{tabular}[c]{@{}l@{}}$n=500,$\\ $\tau=0.95$\end{tabular}} & \multirow{3}{*}{\textsc{SegCertify}} & 0.25 & 0.41 &  & $0.86$ & $0.59$ & $12$ & $31$ & $0.36$ \\
    &  & 0.33 & 0.52 &  & $0.76$ & $0.44$ & $22$ & $48$ & $0.30$ \\
    &  & 0.50 & 0.82 &  & $0.36$ & $0.09$ & $39$ & $51$ & $0.05$ \\
    & \multicolumn{8}{l}{} \\
    & \multirow{3}{*}{\textsc{AdaptiveCertify}} & 0.25 & 0.41 &  & \textbf{0.87 {\fontsize{6}{0}\selectfont\textcolor{ForestGreen}{$1.2\%$}}} & \textbf{0.61 {\fontsize{6}{0}\selectfont\textcolor{ForestGreen}{$3.4\%$}}} & \textbf{9 {\fontsize{6}{0}\selectfont\textcolor{ForestGreen}{$25.0\%$}}} & \textbf{26 {\fontsize{6}{0}\selectfont\textcolor{ForestGreen}{$16.1\%$}}} & $-$ \\
    &  & 0.33 & 0.52 &  & \textbf{0.77 {\fontsize{6}{0}\selectfont\textcolor{ForestGreen}{$1.3\%$}}} & \textbf{0.46 {\fontsize{6}{0}\selectfont\textcolor{ForestGreen}{$4.5\%$}}} & \textbf{18 {\fontsize{6}{0}\selectfont\textcolor{ForestGreen}{$18.2\%$}}} & \textbf{42 {\fontsize{6}{0}\selectfont\textcolor{ForestGreen}{$12.5\%$}}} & $-$ \\
    &  & 0.50 & 0.82 &  & \textbf{0.40 {\fontsize{6}{0}\selectfont\textcolor{ForestGreen}{$11.1\%$}}} & \textbf{0.11 {\fontsize{6}{0}\selectfont\textcolor{ForestGreen}{$22.2\%$}}} & \textbf{28 {\fontsize{6}{0}\selectfont\textcolor{ForestGreen}{$28.2\%$}}} & \textbf{37 {\fontsize{6}{0}\selectfont\textcolor{ForestGreen}{$27.5\%$}}} & $-$ \\
    \bottomrule
    \end{tabularx}
    \caption{Certified segmentation results on the first 200 images in  Cityscapes.}
    \label{tab:cityscapes-table}
    \end{table*}

\begin{table*}[!ht]
    \begin{tabularx}{\textwidth}{lXrrllllll}
    \toprule
    \multicolumn{5}{l}{} & \multicolumn{5}{c}{ACDC} \\
    \cmidrule{6-10}
    &  & \multicolumn{1}{l}{$\sigma$} & \multicolumn{1}{l}{$R$} &  & $\cig\uparrow$ & $c\cig\uparrow$  & $\%\oslash\downarrow$ &  $c\%\oslash\downarrow$ & mIoU $\uparrow$ \\
    \midrule
    & Uncertified HrNet & \multicolumn{1}{l}{-} & \multicolumn{1}{l}{-} &  & $0.61$ & $0.28$ & $-$ & $-$ & $0.15$ \\
    \midrule
    \multirow{7}{*}{\begin{tabular}[c]{@{}l@{}}$n = 100,$\\ $\tau=0.75$\end{tabular}} & \multirow{3}{*}{\textsc{SegCertify}} & $0.25$ & $0.17$ &  & $0.67$ & $0.39$ & $21$ & $37$ & $0.20$ \\
    &  & $0.33$ & $0.22$ &  & $0.57$ & $0.28$ & $27$ & $47$ & $0.17$ \\
    &  & 0.50 & 0.34 &  & $0.25$ & $0.09$ & $26$ & $33$ & $0.04$ \\
    & \multicolumn{8}{l}{} \\
    & \multirow{3}{*}{\textsc{AdaptiveCertify}} & 0.25 & 0.17 &  & \textbf{0.68 {\fontsize{6}{0}\selectfont\textcolor{ForestGreen}{$1.5\%$}}} & \textbf{0.41 {\fontsize{6}{0}\selectfont\textcolor{ForestGreen}{$5.1\%$}}} & \textbf{16 {\fontsize{6}{0}\selectfont\textcolor{ForestGreen}{$23.8\%$}}} & \textbf{29 {\fontsize{6}{0}\selectfont\textcolor{ForestGreen}{$21.6\%$}}} & $-$ \\
    &  & 0.33 & 0.22 &  & \textbf{0.59 {\fontsize{6}{0}\selectfont\textcolor{ForestGreen}{$3.5\%$}}} & \textbf{0.31 {\fontsize{6}{0}\selectfont\textcolor{ForestGreen}{$10.7\%$}}} & \textbf{22 {\fontsize{6}{0}\selectfont\textcolor{ForestGreen}{$18.5\%$}}} & \textbf{35 {\fontsize{6}{0}\selectfont\textcolor{ForestGreen}{$25.5\%$}}} & $-$ \\
    &  & 0.50 & 0.34 &  & \textbf{0.27 {\fontsize{6}{0}\selectfont\textcolor{ForestGreen}{$8.0\%$}}} & \textbf{0.11 {\fontsize{6}{0}\selectfont\textcolor{ForestGreen}{$22.2\%$}}} & \textbf{18 {\fontsize{6}{0}\selectfont\textcolor{ForestGreen}{$30.8\%$}}} & \textbf{19 {\fontsize{6}{0}\selectfont\textcolor{ForestGreen}{$42.4\%$}}} & $-$ \\
    \midrule
    \multirow{7}{*}{\begin{tabular}[c]{@{}l@{}}$n=500,$\\ $\tau=0.95$\end{tabular}} & \multirow{3}{*}{\textsc{SegCertify}} & 0.25 & 0.41 &  & $0.63$ & $0.34$ & $29$ & $51$ & $0.21$ \\
    &  & 0.33 & 0.52 &  & $0.51$ & $0.23$ & $39$ & $63$ & $0.17$ \\
    &  & 0.50 & 0.82 &  & $0.22$ & $0.08$ & $39$ & $49$ & $0.04$ \\
    & \multicolumn{8}{l}{} \\
    & \multirow{3}{*}{\textsc{AdaptiveCertify}} & 0.25 & 0.41 &  & \textbf{0.64 {\fontsize{6}{0}\selectfont\textcolor{ForestGreen}{$1.6\%$}}} & \textbf{0.37 {\fontsize{6}{0}\selectfont\textcolor{ForestGreen}{$8.8\%$}}} & \textbf{25 {\fontsize{6}{0}\selectfont\textcolor{ForestGreen}{$13.8\%$}}} & \textbf{44 {\fontsize{6}{0}\selectfont\textcolor{ForestGreen}{$13.7\%$}}} & $-$ \\
    &  & 0.33 & 0.52 &  & \textbf{0.53 {\fontsize{6}{0}\selectfont\textcolor{ForestGreen}{$3.9\%$}}} & \textbf{0.25 {\fontsize{6}{0}\selectfont\textcolor{ForestGreen}{$8.7\%$}}} & \textbf{34 {\fontsize{6}{0}\selectfont\textcolor{ForestGreen}{$12.8\%$}}} & \textbf{52 {\fontsize{6}{0}\selectfont\textcolor{ForestGreen}{$17.5\%$}}} & $-$ \\
    &  & 0.50 & 0.82 &  & \textbf{0.24 {\fontsize{6}{0}\selectfont\textcolor{ForestGreen}{$9.1\%$}}} & \textbf{0.10 {\fontsize{6}{0}\selectfont\textcolor{ForestGreen}{$25.0\%$}}} & \textbf{31 {\fontsize{6}{0}\selectfont\textcolor{ForestGreen}{$20.5\%$}}} & \textbf{34 {\fontsize{6}{0}\selectfont\textcolor{ForestGreen}{$30.6\%$}}} & $-$ \\
    \bottomrule
    \end{tabularx}
    \caption{Certified segmentation results on the first 200 images in  ACDC.}
    \label{tab:acdc-table}
    \end{table*}

\begin{table*}[!ht]
    \begin{tabularx}{\textwidth}{lXrrllllll}
    \toprule
    \multicolumn{5}{l}{} & \multicolumn{5}{c}{PASCAL-Context} \\
    \cmidrule{6-10}
    &  & \multicolumn{1}{l}{$\sigma$} & \multicolumn{1}{l}{$R$} &  & $\cig\uparrow$ & $c\cig\uparrow$  & $\%\oslash\downarrow$ &  $c\%\oslash\downarrow$ & mIoU $\uparrow$ \\
    \midrule
    & Uncertified HrNet & \multicolumn{1}{l}{-} & \multicolumn{1}{l}{-} &  & $0.58$ & $0.22$ & $-$ & $-$ & $0.15$ \\
    \midrule
    \multirow{7}{*}{\begin{tabular}[c]{@{}l@{}}$n = 100,$\\ $\tau=0.75$\end{tabular}} & \multirow{3}{*}{\textsc{SegCertify}} & $0.25$ & $0.17$ &  & $0.57$ & $0.27$ & $20$ & $27$ & $0.17$ \\
    &  & $0.33$ & $0.22$ &  & $0.46$ & $0.22$ & $31$ & $40$ & $0.14$ \\
    &  & 0.50 & 0.34 &  & $0.15$ & $0.06$ & $41$ & $44$ & $0.03$ \\
    & \multicolumn{8}{l}{} \\
    & \multirow{3}{*}{\textsc{AdaptiveCertify}} & 0.25 & 0.17 &  & \textbf{0.58 {\fontsize{6}{0}\selectfont\textcolor{ForestGreen}{$1.8\%$}}} & \textbf{0.29 {\fontsize{6}{0}\selectfont\textcolor{ForestGreen}{$7.4\%$}}} & \textbf{16 {\fontsize{6}{0}\selectfont\textcolor{ForestGreen}{$20.0\%$}}} & \textbf{21 {\fontsize{6}{0}\selectfont\textcolor{ForestGreen}{$22.2\%$}}} & $-$ \\
    &  & 0.33 & 0.22 &  & \textbf{0.48 {\fontsize{6}{0}\selectfont\textcolor{ForestGreen}{$4.3\%$}}} & \textbf{0.25 {\fontsize{6}{0}\selectfont\textcolor{ForestGreen}{$13.6\%$}}} & \textbf{26 {\fontsize{6}{0}\selectfont\textcolor{ForestGreen}{$16.1\%$}}} & \textbf{32 {\fontsize{6}{0}\selectfont\textcolor{ForestGreen}{$20.0\%$}}} & $-$ \\
    &  & 0.50 & 0.34 &  & \textbf{0.16 {\fontsize{6}{0}\selectfont\textcolor{ForestGreen}{$6.7\%$}}} & \textbf{0.07 {\fontsize{6}{0}\selectfont\textcolor{ForestGreen}{$16.7\%$}}} & \textbf{36 {\fontsize{6}{0}\selectfont\textcolor{ForestGreen}{$12.2\%$}}} & \textbf{39 {\fontsize{6}{0}\selectfont\textcolor{ForestGreen}{$11.4\%$}}} & $-$ \\
    \midrule
    \multirow{7}{*}{\begin{tabular}[c]{@{}l@{}}$n=500,$\\ $\tau=0.95$\end{tabular}} & \multirow{3}{*}{\textsc{SegCertify}} & 0.25 & 0.41 &  & $0.53$ & $0.24$ & $30$ & $41$ & $0.17$ \\
    &  & 0.33 & 0.52 &  & $0.40$ & $0.19$ & $46$ & $57$ & $0.14$ \\
    &  & 0.50 & 0.82 &  & $0.12$ & $0.04$ & $59$ & $61$ & $0.03$ \\
    & \multicolumn{8}{l}{} \\
    & \multirow{3}{*}{\textsc{AdaptiveCertify}} & 0.25 & 0.41 &  & \textbf{0.54 {\fontsize{6}{0}\selectfont\textcolor{ForestGreen}{$1.9\%$}}} & \textbf{0.26 {\fontsize{6}{0}\selectfont\textcolor{ForestGreen}{$8.3\%$}}} & \textbf{26 {\fontsize{6}{0}\selectfont\textcolor{ForestGreen}{$13.3\%$}}} & \textbf{35 {\fontsize{6}{0}\selectfont\textcolor{ForestGreen}{$14.6\%$}}} & $-$ \\
    &  & 0.33 & 0.52 &  & \textbf{0.41 {\fontsize{6}{0}\selectfont\textcolor{ForestGreen}{$2.5\%$}}} & \textbf{0.21 {\fontsize{6}{0}\selectfont\textcolor{ForestGreen}{$10.5\%$}}} & \textbf{41 {\fontsize{6}{0}\selectfont\textcolor{ForestGreen}{$10.9\%$}}} & \textbf{51 {\fontsize{6}{0}\selectfont\textcolor{ForestGreen}{$10.5\%$}}} & $-$ \\
    &  & 0.50 & 0.82 &  & \textbf{0.13 {\fontsize{6}{0}\selectfont\textcolor{ForestGreen}{$8.3\%$}}} & \textbf{0.05 {\fontsize{6}{0}\selectfont\textcolor{ForestGreen}{$25.0\%$}}} & \textbf{55 {\fontsize{6}{0}\selectfont\textcolor{ForestGreen}{$6.8\%$}}} & \textbf{58 {\fontsize{6}{0}\selectfont\textcolor{ForestGreen}{$4.9\%$}}} & $-$ \\
    \bottomrule
    \end{tabularx}
    \caption{Certified segmentation results on the first 200 images in  PASCAL-Context.}
    \label{tab:pascal_ctx-table}
    \end{table*}

\begin{table*}[!ht]
    \begin{tabularx}{\textwidth}{lXrrllllll}
    \toprule
    \multicolumn{5}{l}{} & \multicolumn{5}{c}{COCO-Stuff-10K} \\
    \cmidrule{6-10}
    &  & \multicolumn{1}{l}{$\sigma$} & \multicolumn{1}{l}{$R$} &  & $\cig\uparrow$ & $c\cig\uparrow$  & $\%\oslash\downarrow$ &  $c\%\oslash\downarrow$ & mIoU $\uparrow$ \\
    \midrule
    & Uncertified HrNet & \multicolumn{1}{l}{-} & \multicolumn{1}{l}{-} &  & $0.65$ & $0.36$ & $-$ & $-$ & $0.26$ \\
    \midrule
    \multirow{7}{*}{\begin{tabular}[c]{@{}l@{}}$n = 100,$\\ $\tau=0.75$\end{tabular}} & \multirow{3}{*}{\textsc{SegCertify}} & $0.25$ & $0.17$ &  & $0.58$ & $0.39$ & $20$ & $28$ & $0.25$ \\
    &  & $0.33$ & $0.22$ &  & $0.52$ & $0.32$ & $28$ & $36$ & $0.21$ \\
    &  & 0.50 & 0.34 &  & $0.31$ & $0.15$ & $45$ & $53$ & $0.10$ \\
    & \multicolumn{8}{l}{} \\
    & \multirow{3}{*}{\textsc{AdaptiveCertify}} & 0.25 & 0.17 &  & \textbf{0.60 {\fontsize{6}{0}\selectfont\textcolor{ForestGreen}{$3.4\%$}}} & \textbf{0.41 {\fontsize{6}{0}\selectfont\textcolor{ForestGreen}{$5.1\%$}}} & \textbf{13 {\fontsize{6}{0}\selectfont\textcolor{ForestGreen}{$35.0\%$}}} & \textbf{20 {\fontsize{6}{0}\selectfont\textcolor{ForestGreen}{$28.6\%$}}} & $-$ \\
    &  & 0.33 & 0.22 &  & \textbf{0.54 {\fontsize{6}{0}\selectfont\textcolor{ForestGreen}{$3.8\%$}}} & \textbf{0.34 {\fontsize{6}{0}\selectfont\textcolor{ForestGreen}{$6.3\%$}}} & \textbf{18 {\fontsize{6}{0}\selectfont\textcolor{ForestGreen}{$35.7\%$}}} & \textbf{26 {\fontsize{6}{0}\selectfont\textcolor{ForestGreen}{$27.8\%$}}} & $-$ \\
    &  & 0.50 & 0.34 &  & \textbf{0.35 {\fontsize{6}{0}\selectfont\textcolor{ForestGreen}{$12.9\%$}}} & \textbf{0.18 {\fontsize{6}{0}\selectfont\textcolor{ForestGreen}{$20.0\%$}}} & \textbf{32 {\fontsize{6}{0}\selectfont\textcolor{ForestGreen}{$28.9\%$}}} & \textbf{40 {\fontsize{6}{0}\selectfont\textcolor{ForestGreen}{$24.5\%$}}} & $-$ \\
    \midrule
    \multirow{7}{*}{\begin{tabular}[c]{@{}l@{}}$n=500,$\\ $\tau=0.95$\end{tabular}} & \multirow{3}{*}{\textsc{SegCertify}} & 0.25 & 0.41 &  & $0.53$ & $0.34$ & $31$ & $42$ & $0.24$ \\
    &  & 0.33 & 0.52 &  & $0.45$ & $0.26$ & $41$ & $53$ & $0.20$ \\
    &  & 0.50 & 0.82 &  & $0.26$ & $0.11$ & $61$ & $72$ & $0.09$ \\
    & \multicolumn{8}{l}{} \\
    & \multirow{3}{*}{\textsc{AdaptiveCertify}} & 0.25 & 0.41 &  & \textbf{0.56 {\fontsize{6}{0}\selectfont\textcolor{ForestGreen}{$5.7\%$}}} & \textbf{0.36 {\fontsize{6}{0}\selectfont\textcolor{ForestGreen}{$5.9\%$}}} & \textbf{25 {\fontsize{6}{0}\selectfont\textcolor{ForestGreen}{$19.4\%$}}} & \textbf{35 {\fontsize{6}{0}\selectfont\textcolor{ForestGreen}{$16.7\%$}}} & $-$ \\
    &  & 0.33 & 0.52 &  & \textbf{0.48 {\fontsize{6}{0}\selectfont\textcolor{ForestGreen}{$6.7\%$}}} & \textbf{0.29 {\fontsize{6}{0}\selectfont\textcolor{ForestGreen}{$11.5\%$}}} & \textbf{33 {\fontsize{6}{0}\selectfont\textcolor{ForestGreen}{$19.5\%$}}} & \textbf{44 {\fontsize{6}{0}\selectfont\textcolor{ForestGreen}{$17.0\%$}}} & $-$ \\
    &  & 0.50 & 0.82 &  & \textbf{0.29 {\fontsize{6}{0}\selectfont\textcolor{ForestGreen}{$11.5\%$}}} & \textbf{0.13 {\fontsize{6}{0}\selectfont\textcolor{ForestGreen}{$18.2\%$}}} & \textbf{50 {\fontsize{6}{0}\selectfont\textcolor{ForestGreen}{$18.0\%$}}} & \textbf{62 {\fontsize{6}{0}\selectfont\textcolor{ForestGreen}{$13.9\%$}}} & $-$ \\
    \bottomrule
    \end{tabularx}
    \caption{Certified segmentation results on the first 200 images in  COCO-Stuff-10K.}
    \label{tab:cocostuff-table}
    \end{table*}

\clearpage
\subsection{Visual results}\label{sec:visual}
\begin{figure*}[!ht]
    \captionsetup[subfigure]{labelformat=empty, position=top, justification=centering}
      \subfloat[Input image]
      {\resizebox{.24\textwidth}{!}{\includegraphics{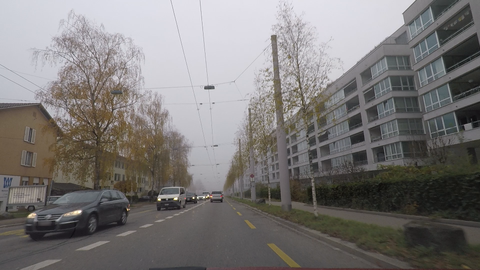}}}
      \hspace{0.25em}
      \subfloat[Ground truth]
      {\resizebox{.24\textwidth}{!}{\includegraphics{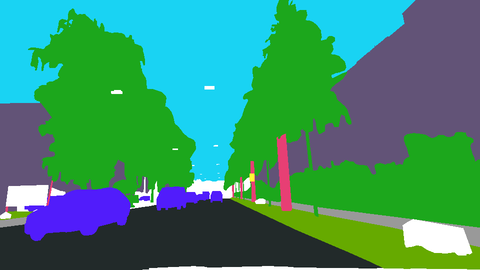}}}
      \hspace{0.25em}
      \subfloat[\textsc{SegCertify} \\$\cig=0.76$,\\$\%\oslash=19.0$]
      {\resizebox{.24\textwidth}{!}{\includegraphics{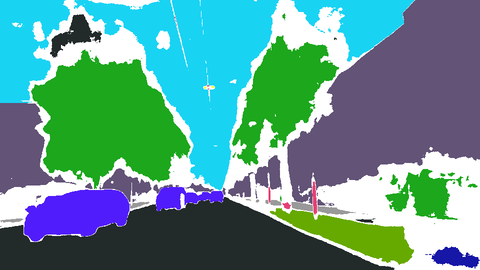}}}
      \hspace{0.25em}
      \subfloat[\textsc{AdaptiveCertify}\\$\cig=0.80 \textcolor{ForestGreen}{(\uparrow5.3\%)}$,\\$\%\oslash=8.2 \textcolor{ForestGreen}{(\downarrow10.9\%)}$]   
      {\resizebox{.24\textwidth}{!}{\includegraphics{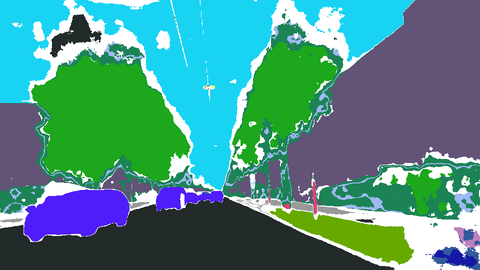}}}
    \\
    \subfloat[]
      {\resizebox{.24\textwidth}{!}{\includegraphics{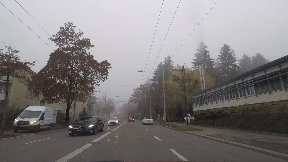}}}
      \hspace{0.25em}
      \subfloat[]
      {\resizebox{.24\textwidth}{!}{\includegraphics{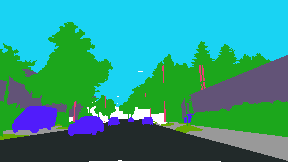}}}
      \hspace{0.25em}
    \subfloat[$\cig=0.82$,\\$\%\oslash=11.7$]
      {\resizebox{.24\textwidth}{!}{\includegraphics{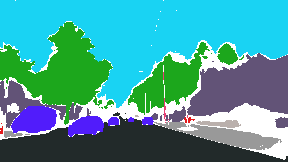}}}
      \hspace{0.25em}
    \subfloat[$\cig=0.84 \textcolor{ForestGreen}{(\uparrow 2.6 \%)}$,\\$\%\oslash=6.6  \textcolor{ForestGreen}{(\downarrow 44.1 \%)}$]   
      {\resizebox{.24\textwidth}{!}{\includegraphics{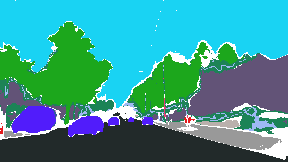}}}
    \\
    \subfloat[]
      {\resizebox{.24\textwidth}{!}{\includegraphics{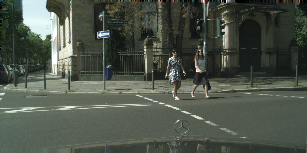}}}
      \hspace{0.25em}
      \subfloat[]
      {\resizebox{.24\textwidth}{!}{\includegraphics{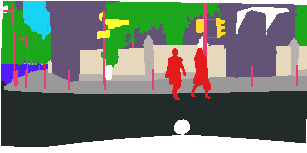}}}
      \hspace{0.25em}
    \subfloat[$\cig=0.79$,\\$\%\oslash=9.9$]
      {\resizebox{.24\textwidth}{!}{\includegraphics{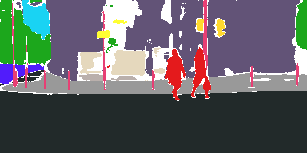}}}
      \hspace{0.25em}
    \subfloat[$\cig=0.81 \textcolor{ForestGreen}{(\uparrow 2.8 \%)}$,\\$\%\oslash=4.9  \textcolor{ForestGreen}{(\downarrow 50.5 \%)}$]   
      {\resizebox{.24\textwidth}{!}{\includegraphics{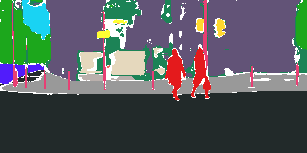}}}
      \\
    \subfloat[]
      {\resizebox{.24\textwidth}{!}{\includegraphics{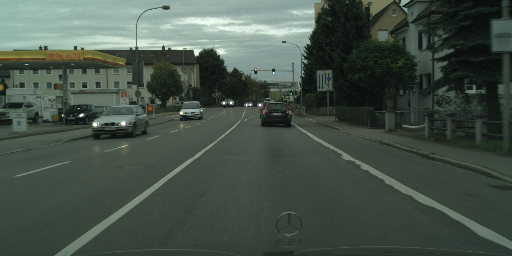}}}
      \hspace{0.25em}
      \subfloat[]
      {\resizebox{.24\textwidth}{!}{\includegraphics{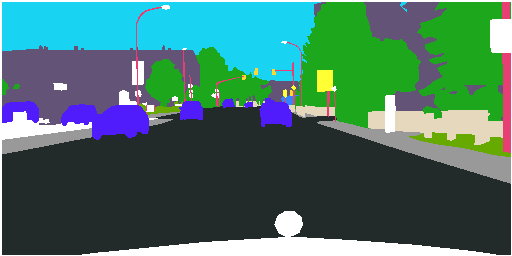}}}
      \hspace{0.25em}
    \subfloat[$\cig=0.87$,\\$\%\oslash=10.8$]
      {\resizebox{.24\textwidth}{!}{\includegraphics{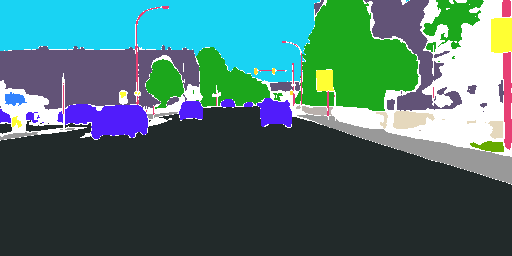}}}
      \hspace{0.25em}
    \subfloat[$\cig=0.89 \textcolor{ForestGreen}{(\uparrow 2.9 \%)}$,\\$\%\oslash=4.9  \textcolor{ForestGreen}{(\downarrow 54.3 \%)}$]   
      {\resizebox{.24\textwidth}{!}{\includegraphics{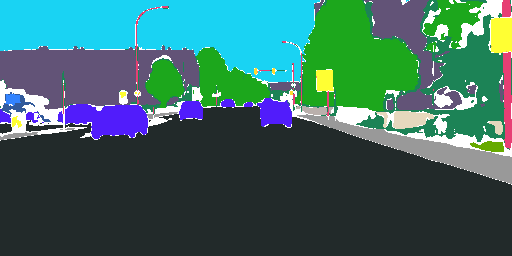}}}
    
    \caption{Selected examples showcasing the performance of \textsc{AdaptiveCertify} against \textsc{SegCertify}. }
\label{fig:handpicked-visual}
\end{figure*}

In prior sections, we examined the quantitative performance of adaptive hierarchical certification compared to the non-adaptive baseline using the $\cig$ and abstention rate $\%\oslash$ metrics. In this section, we look more into how the certified segmentation output looks like qualitatively by both methods. We sample images from both datasets Cityscapes and ACDC to visually evaluate the performance of \textsc{AdaptiveCertify} (ours) against \textsc{SegCertify}. In Figure \ref{fig:handpicked-visual}, we show selected examples that resemble a significant improvement by \textsc{AdaptiveCeritify} against the baseline in terms of the certified information gain ($\cig$) and abstain rate ($\%\oslash$). 

\clearpage
\subsection{Boundary pixels analysis (Extended)}\label{subsection:boundary}

In prior sections, we demonstrated that \textsc{AdaptiveCertify} enhances the certification performance, quantifiable by an increase in $\cig$, $c\cig$ and reduction in the abstention rate $\oslash\%$ and class-averaged abstention rate $c\oslash\%$, relative to the baseline method. Given that initial visual observations from Section \ref{sec:visual} suggest that the abstention predominantly occurs at object boundaries, a detailed examination was necessary. We segmented the images into boundary and non-boundary pixels to systematically analyze the abstention behavior on these pixels by both \textsc{AdaptiveCertify} and the baseline. This analysis aims to elucidate the specific areas where the baseline abstains, as well as show the performance of \textsc{AdaptiveCertify} across both pixel types. 

\begin{figure*}[!ht]
    \captionsetup[subfigure]{justification=centering}
    \centering
    \subfloat[Ground truth]
    {\resizebox{0.32\textwidth}{!}{\includegraphics{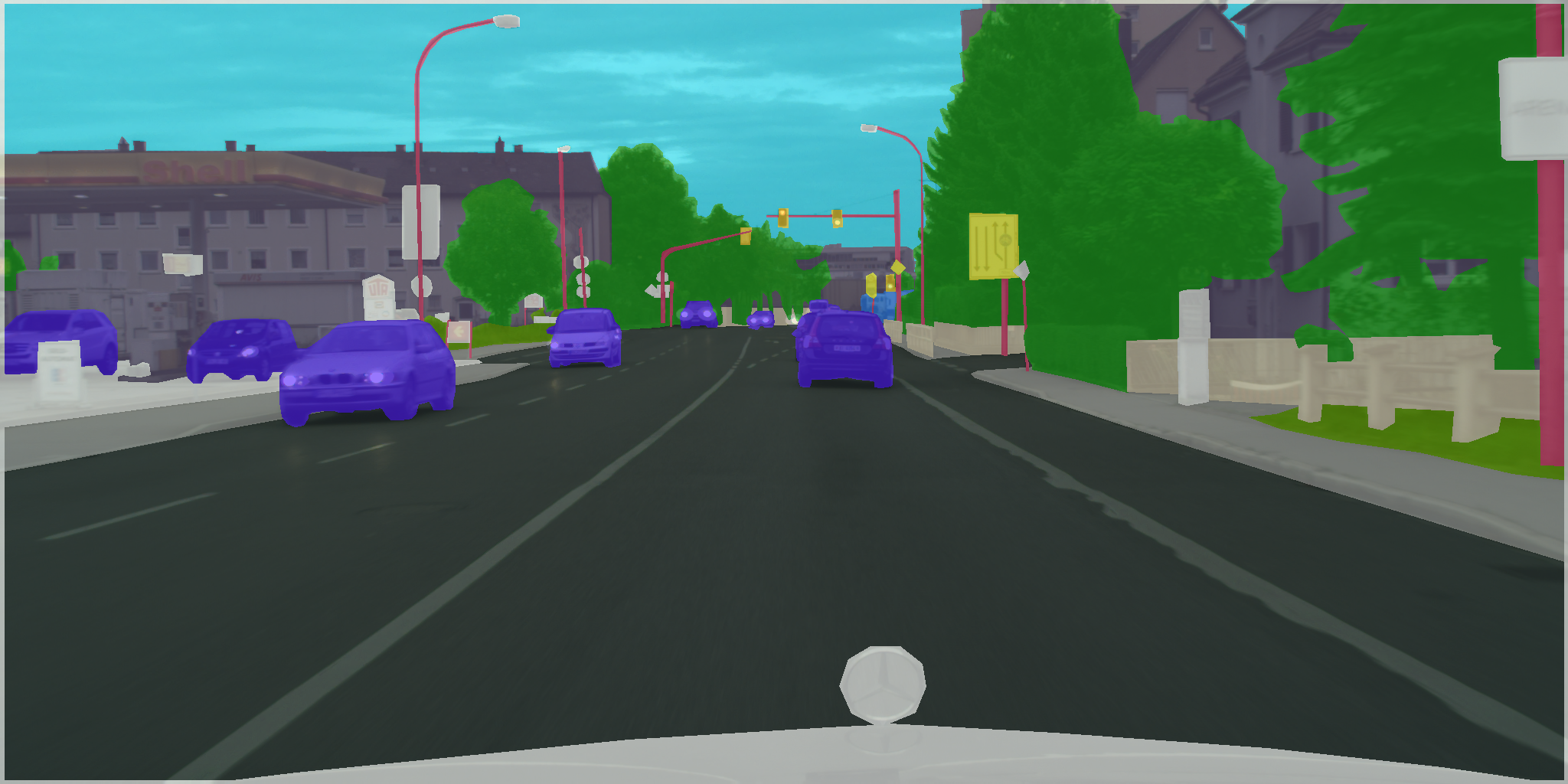}}} 
    \hspace{0.5em}
    \subfloat[Certified segmentation \\ (Baseline)\\ $\cig=0.87$,\\ $\%\oslash=10.8$]  
    {\resizebox{0.32\textwidth}{!}{\includegraphics{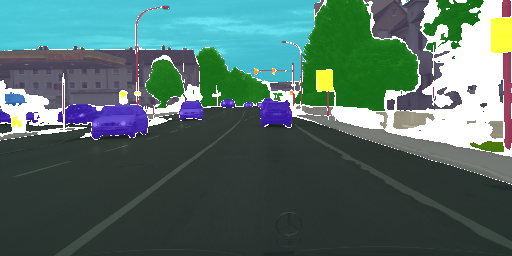}}} 
    \hspace{0.5em}
    \subfloat[Certified Segmentation (\textsc{AdaptiveCertify})\\ $\cig=0.89 \textcolor{ForestGreen}{(\uparrow 2.9 \%)}$,\\$\%\oslash= 4.9  \textcolor{ForestGreen}{(\downarrow 54.3 \%)}$]  
    {\resizebox{0.32\textwidth}{!}{\includegraphics{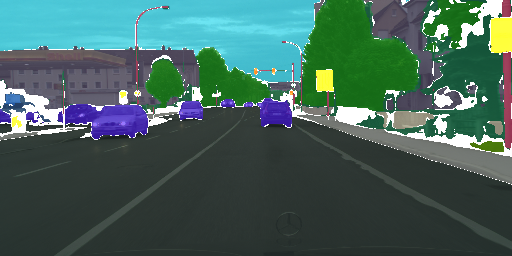}}} 
    \\
    \subfloat[Boundary map]  
    {\resizebox{0.32\textwidth}{!}{\includegraphics{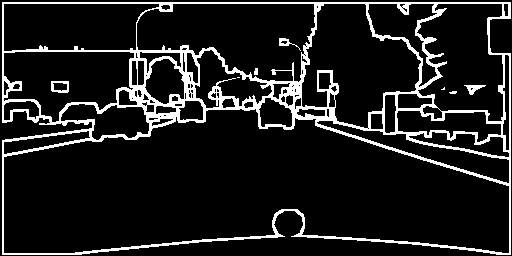}}}
    \hspace{0.5em}
    \subfloat[$\oslash \cap$ boundary (Baseline)\\ $\oslash \% \text{boundary}=21.0$]  
    {\resizebox{0.32\textwidth}{!}{\includegraphics{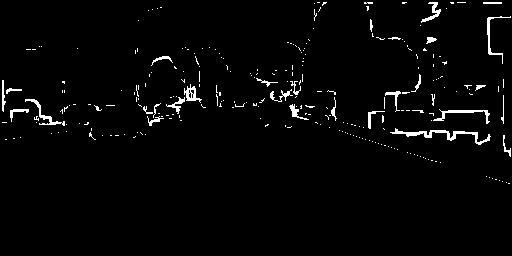}}}
    \hspace{0.5em}
    \subfloat[$\oslash \cap$boundary (\textsc{AdaptiveCertify})\\ $\oslash \% \text{boundary}=11.9  \textcolor{ForestGreen}{(\downarrow 43.1 \%)}$]  
    {\resizebox{0.32\textwidth}{!}{\includegraphics{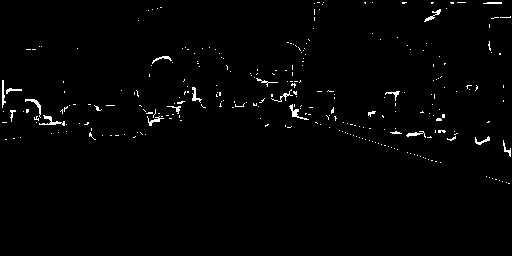}}}
    \\
    \subfloat[Non-boundary map]  
    {\resizebox{0.32\textwidth}{!}{\includegraphics{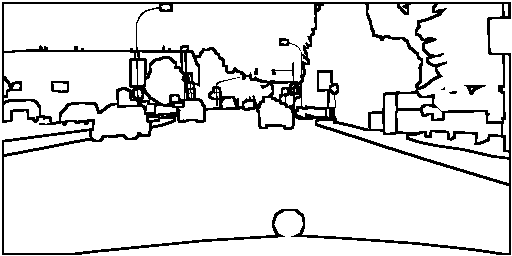}}}
    \hspace{0.5em}
    \subfloat[$\oslash \cap$non-boundary (Baseline)\\ $\oslash \% \text{non-boundary}=9.3$]  
    {\resizebox{0.32\textwidth}{!}{\includegraphics{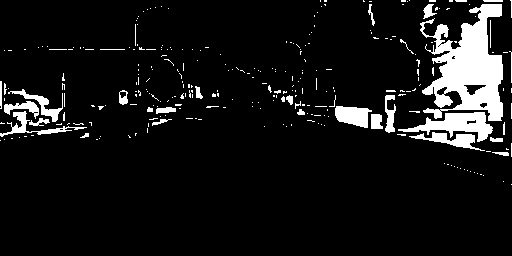}}}
    \hspace{0.5em}
    \subfloat[$\oslash \cap$non-boundary (\textsc{AdaptiveCertify})\\ $\oslash \% \text{boundary}=3.9  \textcolor{ForestGreen}{(\downarrow 58.1 \%)}$]  
    {\resizebox{0.32\textwidth}{!}{\includegraphics{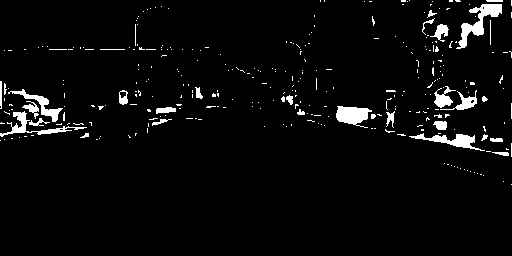}}}
    \\
    \caption{A visual example showing how abstain pixels by the baseline (\textsc{SegCertify} and our method \textsc{AdaptiveCertify} intersect with boundary and non-boundary pixels. The first row shows the ground truth segmentation map (a) and the certified segmentation outputs by the baseline and ours in (b) and (c). The second row shows the boundary map (d) extracted from the groud truth map in (a), and the map representing the intersection between baseline and our abstain $\oslash$ pixels and boundary pixels in (e) and (f). The third row follows the same second row logic, except for non-boundary pixels.}
    \label{fig:b-analysis}
    \end{figure*}

To conduct this analysis, we need to differentiate between boundary and non-boundary pixels given a ground truth segmentation map. The way we isolate boundary pixels is by first applying 2D convolution over the ground truth segmentation map with a kernel that isolates central pixels within a defined area (controlled by the margin parameter we set to $10$), followed by a dilation process that expands these central values over the surrounding pixels. The function then compares the isolated central values to the dilated values: where these differ, boundary pixels are identified, highlighting transitions between different segmentation labels. The non-boundary map is intuitively the complement of the boundary map generated via this process. An example of a boundary map is shown in Figure \ref{fig:b-analysis} (d) which is extracted from the ground truth segmentation map in (a).

\paragraph{Qualitative analysis} In Figure \ref{fig:b-analysis}, we examine the distribution of abstained pixels relative to boundary and non-boundary regions in a visual example from the Cityscapes dataset. The ground truth map displayed in (a) provides the reference for actual image boundaries, aiding in the assessment of certification outputs by \textsc{SegCertify} (baseline) and \textsc{AdaptiveCertify} (our method), shown in (a) and (b) respectively. A significant reduction in boundary pixel abstention by our method is evident in (f). Additionally, the third row features abstention analysis for non-boundary pixels, where \textsc{AdaptiveCertify} also demonstrates a substantial decrease in the abstention of non-boundary pixels compared to the baseline.

\subsection{Per-class Performance (Extended)}\label{subsection:delta}
\begin{figure}[!ht]
\centering
\resizebox{0.7\textwidth}{!}{\includegraphics[]{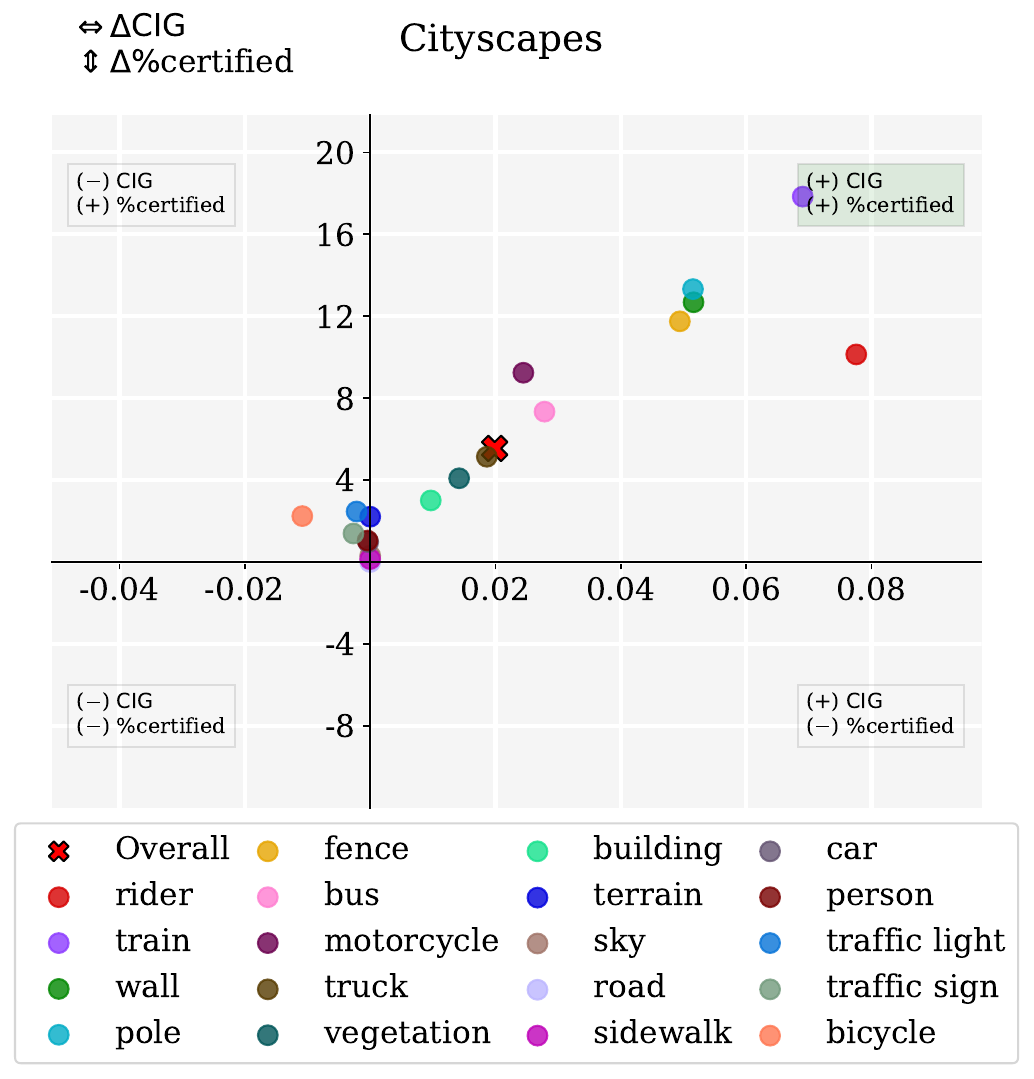}}  
\caption{The performance of \textsc{AdaptiveCertify} against the baseline in terms of the difference in $\cig$ ($\Delta \cig$) and the certification rate ($\Delta \%$certified) on all classes in the Cityscapes dataset. "Overall" indicates the class-average performance. The top right quadrant indicates that \textsc{AdaptiveCertify} outperforms the baseline in both metrics. The results are averaged over the first $100$ images in the dataset.}
\label{fig:delta-cs}
\end{figure}

\begin{figure}[!ht]
\centering
\resizebox{0.7\textwidth}{!}{\includegraphics[]{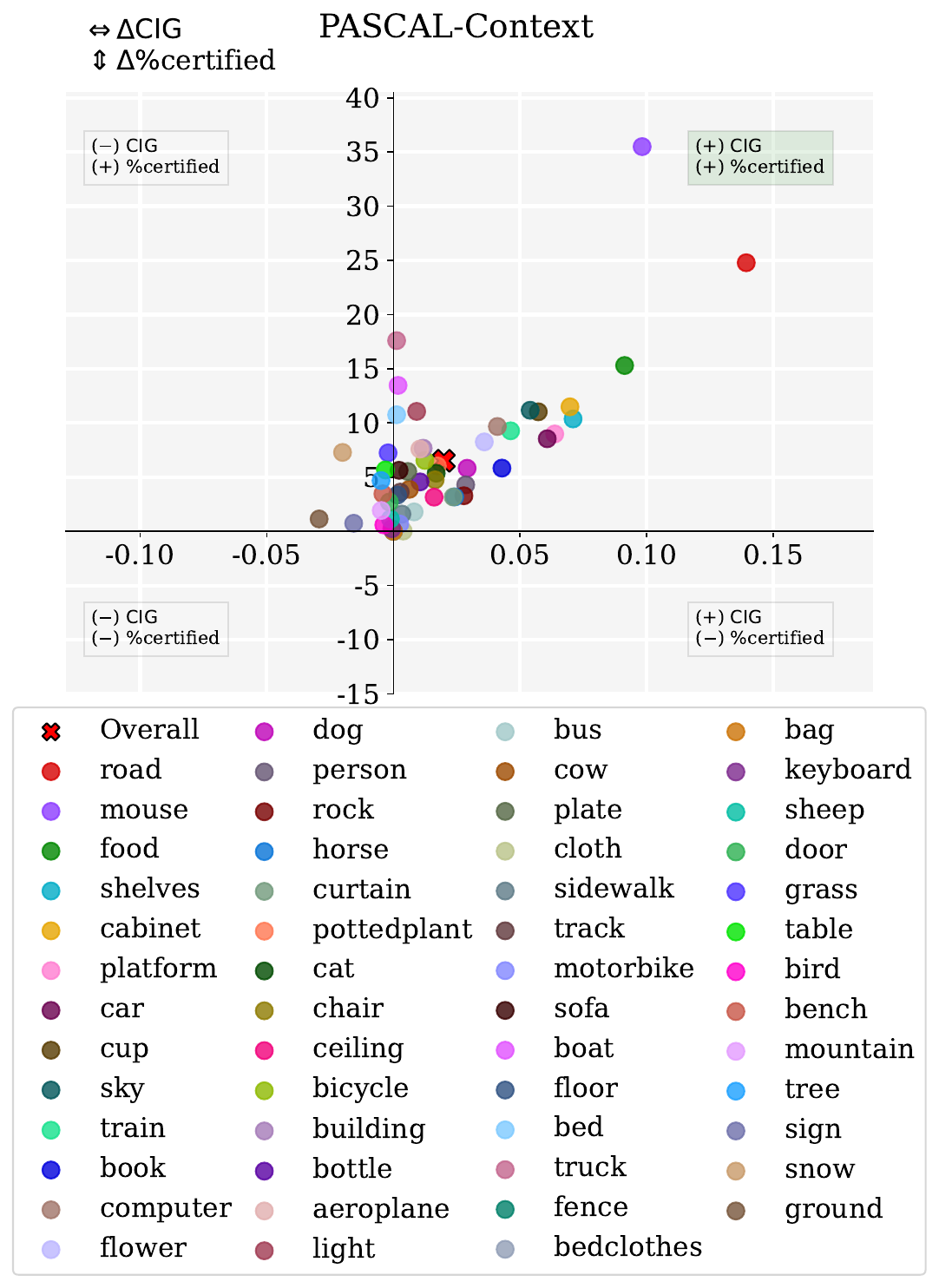}}  
\caption{The performance of \textsc{AdaptiveCertify} against the baseline in terms of the difference in $\cig$ ($\Delta \cig$) and the certification rate ($\Delta \%$certified) on all classes in the PASCAL-Context dataset. "Overall" indicates the class-average performance. The top right quadrant indicates that \textsc{AdaptiveCertify} outperforms the baseline in both metrics. The results are averaged over the first $100$ images in the dataset.}
\label{fig:pascal-delta-ext}
\end{figure}

\begin{figure}[!ht]
\centering
\resizebox{0.6\textwidth}{!}{\includegraphics[]{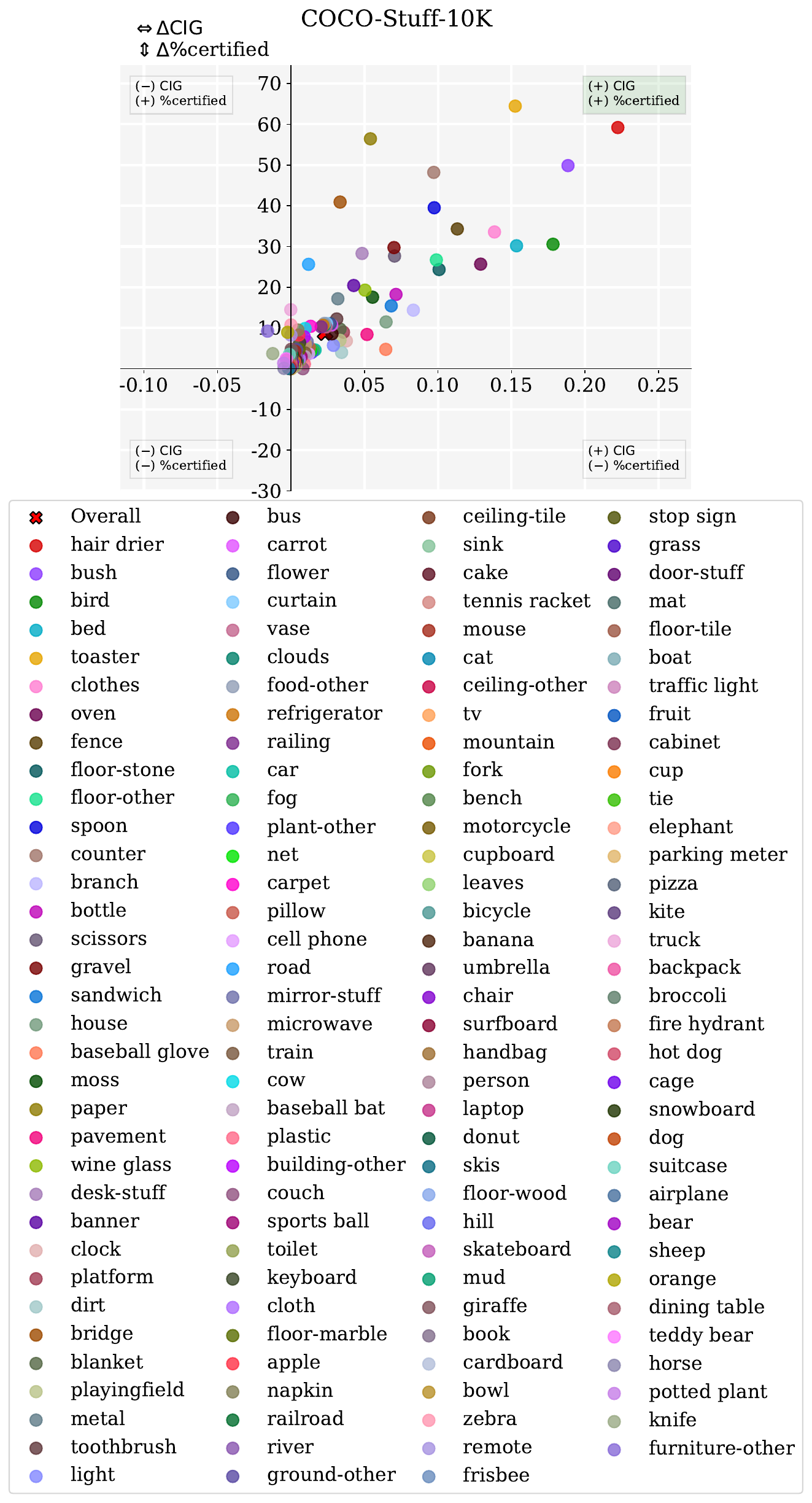}}  
\caption{The performance of \textsc{AdaptiveCertify} against the baseline in terms of the difference in $\cig$ ($\Delta \cig$) and the certification rate ($\Delta \%$certified) on all classes in the COCO-Stuff-10K dataset. "Overall" indicates the class-average performance. The top right quadrant indicates that \textsc{AdaptiveCertify} outperforms the baseline in both metrics. The results are averaged over the first $100$ images in the dataset.}
\label{fig:coco-delta-ext}
\end{figure}

An extension of Figure \ref{fig:delta-all} for all classes in Cityscapes, PASCAL-Context and and COCO-Stuff-10K datasets is in Figures \ref{fig:delta-cs}, \ref{fig:pascal-delta-ext} and \ref{fig:coco-delta-ext}, showing the per-class performance of our adaptive hierarchical method \textsc{AdaptiveCertify} against the baseline \textsc{SegCertify}.


\clearpage

\subsection{CIG and Abstention Rate Tradeoff}\label{subsection:tradeoff}

The core idea behind adaptive hierarchical certification is to group the fluctuating classes in unstable components under a higher concept that contains them within a pre-defined semantic hierarchy. To maximize the certification rate (and minimize the abstention rate), it would be trivial, however, to group all classes under a single vertex at the most coarse level in a semantic hierarchy. This way, the certification rate will be maximized, at the cost of also having a low Certified Information Gain. On the other hand, sticking to the most fine-grained level in the hierarchy $H_0$ imposes a conservative requirement that causes low certification rates, as seen in \textsc{SegCertify}. Motivated by these two extremes—either sampling from the most coarse or fine-grained levels in the hierarchy—we look more into how adaptive hierarchical certification performs against such non-adaptive baselines.

\begin{figure}[!ht]
\subfloat[]
{\resizebox{\textwidth}{!}{\includegraphics[]{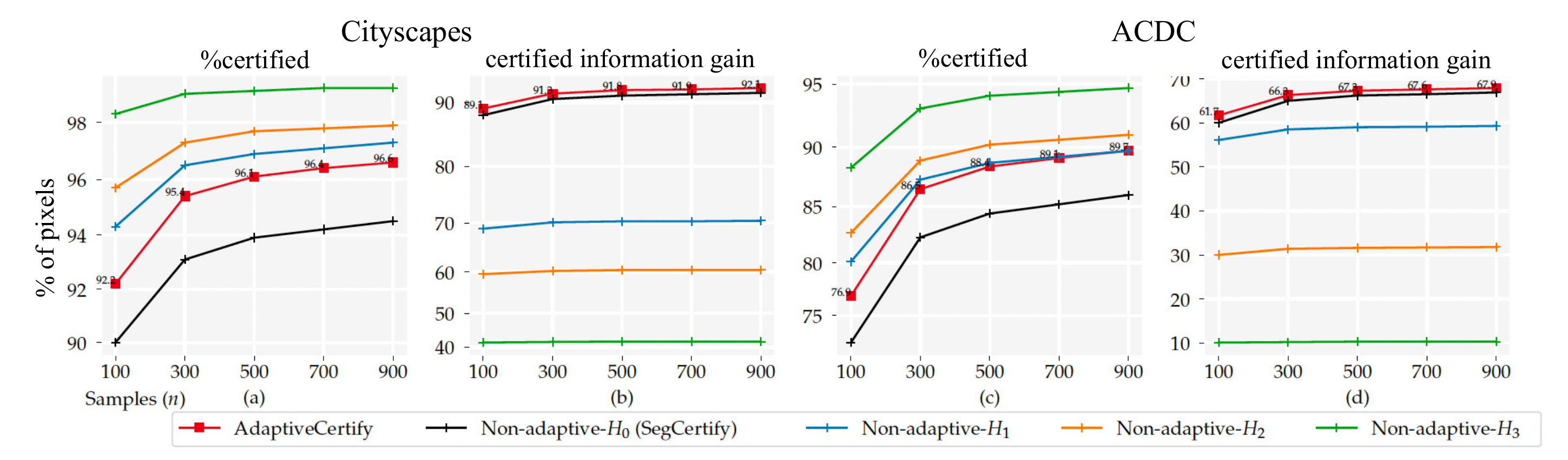}}}
\caption{\%certified (mean per-pixel certification rate) and Certified Information Gain versus the number of samples ($n$) on Cityscapes and ACDC.}
\label{fig:tradeoff}
\end{figure}

To do so, we investigate the performance of additional baselines on Cityscapes and ACDC, where each of them samples from a different level of the hierarchy, starting from the most fine-grained $H_0$ (Non-adaptive-$H_0$ is \textsc{SegCertify}) up to the most coarse $H_3$, namely Non-adaptive-$H_3$, in Figure \ref{fig:tradeoff}. We particularly want to investigate the relationship between the Certified Information Gain ($\cig$) and certification rate (\%certified).

In Figure \ref{fig:tradeoff}, generally, for the non-adaptive baselines, the higher the level is, the higher the certification rate and lower the Certified Information Gain, and vice versa. We notice that the certification rate is the lowest in Non-adaptive-$H_0$ across different numbers of samples ($n$). Meanwhile, it is the highest for Non-adaptive-$H_3$, since $H_3$ is the most coarse level in the Cityscapes and ACDC hierarchy (refer to Figure \ref{fig:cs-h} for the hierarchy). The higher the hierarchy level, the less information is retained due to the grouping of more and more classes. On the other hand, more pixels can be certified, as the number of fluctuating components decreases. By comparing the performance of the adaptive method \textsc{AdaptiveCertify} and the other non-adaptive baselines, we notice two things: \textsc{AdaptiveCertify} maintains the highest $\cig$ across different $n$ and $\sigma$, while it has a higher certification rate compared to Non-adaptive-$H_0$, but lower than the rest with $H_i$ such that $i>0$. Our hierarchical approach combines the best of both worlds, retaining the most amount of information from lower hierarchy levels where possible and falling back to higher hierarchy levels to avoid abstaining.

\subsection{Definition: Generality of a Vertex}
The generality of the vertex $v_i$ denoted by $\mathrm{G}(v_i)$ is defined as the number of leaf vertices that are reachable by $v_i$, in other words, the number of leaf descendants of $v_i$. $G$ is formally described as
\begin{equation}\label{eq:gen}
    G(v_i) = | \{ y_j \mid \left(\exists_{v_i, v_{i-1} \ldots, v_1, y_j} \{(v_i, v_{i-1}), \ldots, (v_1, y_j)\} \subseteq \mathcal{E} \right) \land \left( y_j \in H_0 \right) \} |
\end{equation}

where it is the carnality of the set of leaf vertices $y_j \in H_0$ such that there is a path from the vertex $v_i$ to $y_j$.

\end{document}